\documentclass[acmtog]{acmart}
\acmSubmissionID{1975}

\usepackage{booktabs} 

\usepackage[ruled]{algorithm2e} 

\SetAlFnt{\small}
\SetAlCapFnt{\small}
\SetAlCapNameFnt{\small}
\SetAlCapHSkip{0pt}

\usepackage{xcolor}
\usepackage{pifont}
\newcommand{\tablestyle}[2]{\setlength{\tabcolsep}{#1}\renewcommand{\arraystretch}{#2}\centering}
\definecolor{gold}{HTML}{FAE37F}
\definecolor{silver}{HTML}{D7D7D7}
\definecolor{bronze}{HTML}{EDBA91}
\newcommand{\au}[1]{\setlength{\fboxsep}{1pt}\colorbox{gold}{#1}}
\newcommand{\ag}[1]{\setlength{\fboxsep}{1pt}\colorbox{silver}{#1}}

\newcommand{\gcheck}[0]{\color{green}\ding{52}}
\newcommand{\rcross}[0]{\color{red}\ding{56}}

\definecolor{modcolor}{rgb}{0,0,0.7}

\acmJournal{TOG}
\copyrightyear{2025}
\acmYear{2025}
\setcopyright{acmlicensed}\acmConference[SA Conference Papers '25]{SIGGRAPH Asia 2025 Conference Papers}{December 15--18, 2025}{Hong Kong, Hong Kong}
\acmBooktitle{SIGGRAPH Asia 2025 Conference Papers (SA Conference Papers '25), December 15--18, 2025, Hong Kong, Hong Kong}
\acmDOI{10.1145/3757377.3763955}
\acmISBN{979-8-4007-2137-3/2025/12}

\begin{document}
\title{ARTalk: Speech-Driven 3D Head Animation via Autoregressive Model}

\author{Xuangeng Chu}
\affiliation{
    \institution{The University of Tokyo}
    \city{Tokyo}
    \country{Japan}
}
\email{xuangeng.chu@mi.t.u-tokyo.ac.jp}

\author{Nabarun Goswami}
\affiliation{
    \institution{The University of Tokyo}
    \city{Tokyo}
    \country{Japan}
}
\email{nabarungoswami@mi.t.u-tokyo.ac.jp}

\author{Ziteng Cui}
\affiliation{
    \institution{The University of Tokyo}
    \city{Tokyo}
    \country{Japan}
}
\email{cui@mi.t.u-tokyo.ac.jp}

\author{Hanqin Wang}
\affiliation{
    \institution{The University of Tokyo}
    \city{Tokyo}
    \country{Japan}
}
\email{wang@mi.t.u-tokyo.ac.jp}

\author{Tatsuya Harada}
\affiliation{
    \institution{The University of Tokyo}
    \institution{, RIKEN AIP}
    \city{Tokyo}
    \country{Japan}
}
\email{harada@mi.t.u-tokyo.ac.jp}

\begin{abstract}
Speech-driven 3D facial animation aims to generate realistic lip movements and facial expressions for 3D head models from arbitrary audio clips.
Although existing diffusion-based methods are capable of producing natural motions, their slow generation speed limits their application potential.
In this paper, we introduce a novel autoregressive model that achieves real-time generation of highly synchronized lip movements and realistic head poses and eye blinks by learning a mapping from speech to a multi-scale motion codebook.
Furthermore, our model can adapt to unseen speaking styles, enabling the creation of 3D talking avatars with unique personal styles beyond the identities seen during training. 
Extensive evaluations and user studies demonstrate that our method outperforms existing approaches in lip synchronization accuracy and perceived quality.
\end{abstract}

%
%

%
%


\maketitle

\section{Introduction}

\begin{figure}[ht]
\begin{center}
\centerline{\includegraphics[width=1.0\columnwidth]{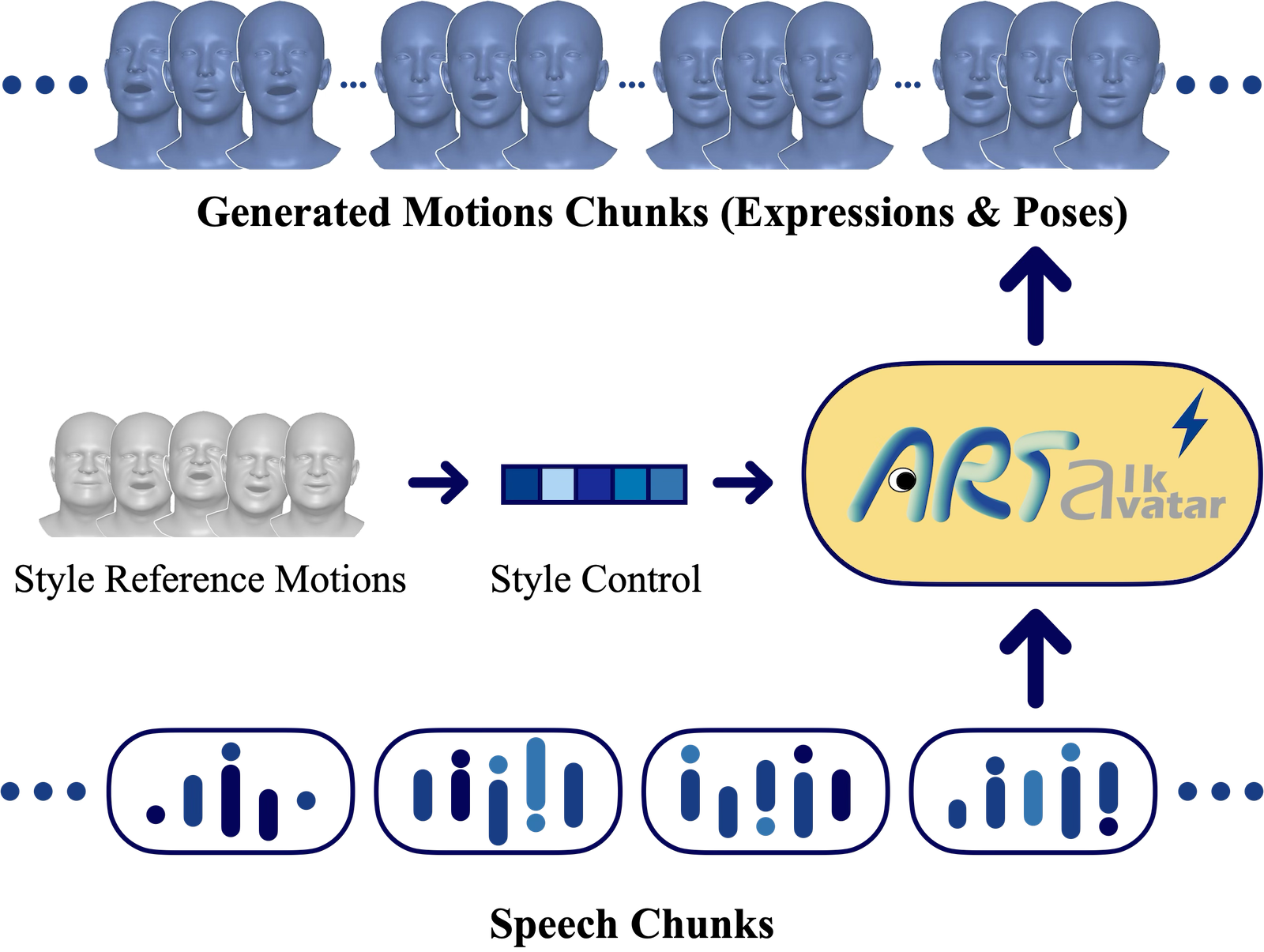}}
\caption{
We present ARTalk, a framework for speech-driven 3D facial motion generation.
Our method learns a mapping from speech to a multi-scale motion code, enabling the real-time generation of realistic and diverse animation sequences.
}
\label{fig:teaser}
\end{center}
\Description{A concept pipeline of our method.}
\end{figure}

The rapid evolution of large language models \citep{achiam2023gpt,team2023gemini,guo2025deepseek} in recent years has driven significant progress in artificial intelligence, particularly in generating human-like text and facilitating complex conversational interactions.
However, many such systems primarily rely on text or synthesized speech for communication, often lacking the visual embodiment necessary for achieving rich and intuitive human-computer interaction.
To bridge this gap and create more engaging interactive experiences, research into digital humans has garnered widespread attention from both academia and industry.
Due to the intrinsic and natural correlation between the speech and facial expressions, speech-driven 3D facial animation has emerged as a key component for shaping lifelike virtual characters.
Therefore, advancements in this field are essential for promoting the application of digital humans in various scenarios such as education and entertainment.

Speech-driven 3D head motion generation aims to synthesize natural and temporally synchronized facial expressions, particularly lip movements, along with realistic head motions directly from speech input.
Recent years have witnessed significant advancements in speech-driven motion generation.
Autoregressive models, for instance, have been explored to predict deformation offsets on 3D meshes \citep{richard2021meshtalk, codetalker2023, multitalk2024, scantalk2024}.
However, the inherent complexity of representing facial motion through mesh deformations in 3D space often struggles to fully capture the intricate and non-linear mappings between speech and the resulting nuanced facial expressions.
This limitation frequently leads to over-smoothed non-lip regions and a lack of fine-grained realism.
More recently, diffusion models \citep{ho2020denoising, tevet2023human, facediffuser2023, diffposetalk2024} have demonstrated compelling generative capabilities, including in the domain of speech-driven motion.
For example, FaceDiffuser \citep{facediffuser2023} leverages diffusion processes to generate motion directly on 3D meshes, while DiffPoseTalk \citep{diffposetalk2024} utilizes diffusion on the parametric FLAME model \citep{FLAME2017} blendshapes to produce natural facial expressions and head movements.
Despite their impressive results in generating high-fidelity motions, these diffusion-based approaches typically suffer from high computational costs, hindering their applicability in real-time and low-latency scenarios.
In parallel, multi-scale autoregressive models \citep{var2024} have achieved notable success in image generation tasks, offering a balance between generation speed and output quality.
However, these architectures are primarily designed for processing single, multi-scale static samples rather than sequential temporal data, making their direct application to the temporal dynamics inherent in speech-to-motion generation challenging.

To overcome the limitations of existing methods and achieve high-quality, natural, and fast speech-driven motion generation, we propose a novel approach based on temporal windows and multi-scale autoregression.
Our method segments the speech input into consecutive temporal windows for motion generation, a window-wise processing strategy crucial for achieving low-latency motion synthesis.
Within each temporal window, we further employ a multi-scale encoding and generation strategy to capture facial expressions and head movements with rich details.
For instance, for a 4-second temporal window containing 100 motion frames, we encode it into a set of multi-scale representations with varying temporal resolutions, such as encodings with lengths of [1, 5, 25, 50, 100], and adopt a similar multi-scale decoding process during generation. This design allows the model to simultaneously capture global temporal context and fine-grained details of each motion frame, resulting in more expressive motions.
To minimize encoding errors and ensure temporal coherence of the generated motions, we extend a VQ codec to encode multi-scale motion codes across two consecutive temporal windows, thereby effectively modeling and transferring temporal dependencies between windows.
Additionally, we design a novel autoregressive generator that can autoregressively predict the multi-scale motion codes for the current temporal window conditioned on the speech features of the current window and the previously generated motion sequence.
This conditional generation mechanism enables our model to achieve high-quality multi-scale motion generation within each temporal window while maintaining smooth transitions and temporal consistency across windows.
Furthermore, we integrate a Transformer-based style encoder to extract style representations from example motion clips. This allows our model to generate stylized motions with unique individual characteristics and helps to disentangle the complex many-to-many mapping between speech and motion.
Through these innovations, our method can generate personalized and high-fidelity lip movements, facial expressions, and head motions, demonstrating significant potential for various speech-driven downstream applications.

Our main contributions are as follows:
\begin{itemize}
\item We propose ARTalk, a novel autoregressive framework capable of generating natural 3D facial motions with head poses in real time.
\item We design a novel encoder-decoder that encode motions from consecutive time windows to produce temporally dependent multi-scale motion representations.
\item We introduce a novel conditional autoregressive generator on temporal and feature scales, enabling motion generations tightly aligned with speech conditions and consistent across time windows.
\end{itemize}

\section{Related Work}
\begin{table}[t]
\caption{
    Comparison across methods, including style adaptation during inference (w/ Style), head pose generation (w/ Pose), and real-time capability (Real-time).
    ARTalk (ours) is the only method to achieve all three features, demonstrating its comprehensive advantages over baseline methods.
}
\label{tab:comp_methods}
\begin{center}
\tablestyle{5pt}{1.05}
\scalebox{0.95}{
    \begin{tabular}{lccc}
        \toprule
        Method & w/ Style & w/ Pose & Real-time \\
        \midrule
        FaceFormer \citep{faceformer2022}       & \rcross & \rcross & \rcross  \\
        CodeTalker \citep{codetalker2023}       & \rcross & \rcross & \rcross  \\
        SelfTalk \citep{selftalk2023}           & \rcross & \rcross & \gcheck  \\
        FaceDiffuser \citep{facediffuser2023}   & \rcross & \rcross & \rcross  \\
        MultiTalk \citep{multitalk2024}         & \gcheck & \rcross & \rcross  \\
        ScanTalk \citep{scantalk2024}           & \rcross & \rcross & \rcross  \\
        UniTalker \citep{fan2024unitalker}      & \gcheck & \rcross & \rcross  \\
        DiffPoseTalk \citep{diffposetalk2024}   & \gcheck & \gcheck & \rcross  \\
        ARTalk (Ours)                           & \gcheck & \gcheck & \gcheck \\
        \bottomrule
    \end{tabular}
}
\end{center}
\end{table}

\begin{figure*}[t]
\begin{center}
\centerline{\includegraphics[width=1.0\linewidth]{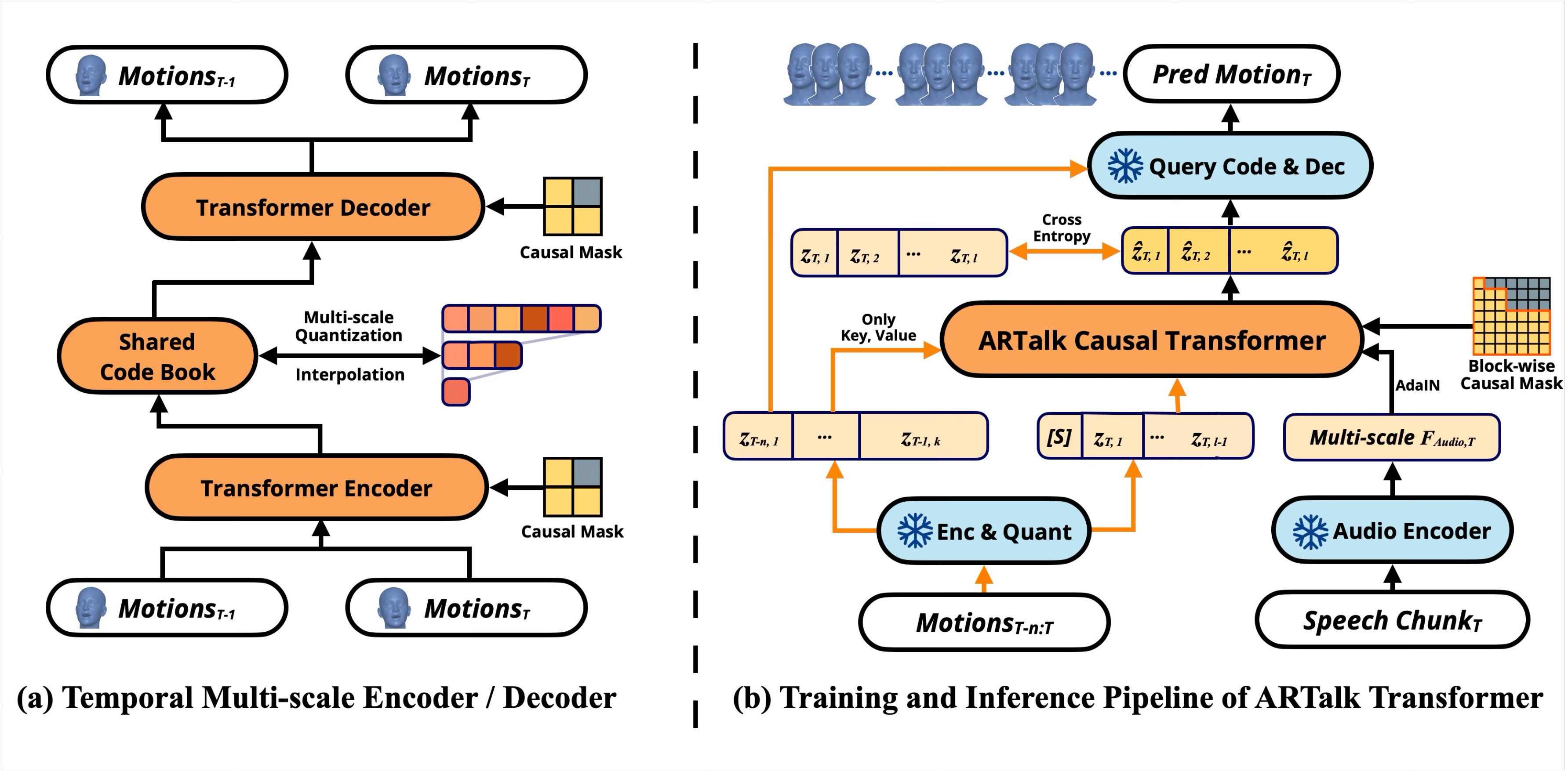}}
\vskip -0.5cm
\caption{
ARTalk involves two separated parts. (a) shows our temporal multi-scale VQ autoencoder. It encodes motion sequences into multi-scale token maps \([M_{k_1}, M_{k_2}, ..., M_{K}]\) using a shared codebook and causal masking on temporal. (b) shows The ARTalk Causal Transformer, where training uses ground truth tokens with a block-wise causal attention mask, and inference autoregressively predicts motion tokens conditioned on speech features and last scale tokens and last time window motions.
}
\label{fig:main_method}
\end{center}
\vskip -0.1cm
\Description{Method pipeline of our method.}
\end{figure*}

\subsection{Speech-Driven 3D Facial Animation}
Research on audio-driven 3D motion generation has been an area of interest for decades, with methods evolving significantly over time.
Early approaches \citep{taylor2012dynamic, xu2013practical,edwards2016jali} relied primarily on procedural methods, which segmented speech into phonemes and mapped them to predefined visemes through handcrafted rules.
Although procedural methods provide explicit control over the generated animations, they typically require complex parameter tuning and fail to capture the diversity and complexity of real-world speaking styles.
As a result, procedural methods struggled to deliver animations that appeared both natural and adaptable to varying speech dynamics.
In recent years, learning-based approaches \citep{faceformer2022, codetalker2023, diffposetalk2024, danvevcek2023emotional, emotalk2023, yang2024probabilistic, fan2024unitalker, chae2025perceptually} have advanced rapidly, addressing these limitations and enabling more natural and expressive facial animations.
Additionally, some approaches \citep{yi2022predicting, ye2024real3d, zhang2023sadtalker, tan2024say} focus on directly generating talking head videos without explicitly modeling motion.
However, this limits their ability to integrate with motion-driven downstream applications, restricting their broader applicability.

\subsection{Autoregressive 3D Facial Motion Generation}
Autoregressive (AR) methods model the temporal sequence of facial motion in a step-by-step manner.
These methods typically leverage pre-trained speech model features \citep{baevski2020wav2vec, hsu2021hubert, kyutai2024moshi} as speech representations, which are subsequently mapped to 3D deformable model parameters or 3D meshes via neural networks.
For example, FaceFormer \citep{faceformer2022} autoregressively predicts the continuous facial motion parameters while MeshTalk \citep{richard2021meshtalk} learns a prior over categorical expression tokens.
CodeTalker \citep{codetalker2023} and MultiTalk \citep{multitalk2024} tokenize facial motions with a VQ-VAE and train a transformer decoder to predict subsequent motion tokens based on audio features.
Learn2Talk \citep{Zhuang2024learn2talk} augments FaceFormer with lip-sync and lip-reading losses, improving alignment with speech.
MMHead \citep{wu2024mmhead} provides text descriptions as an extra condition for controlling the generated motion. 
However, these methods also show limitations.
Temporal autoregressive modeling often under represents motion within each time window, leading to overly smooth lip movements and failing to capture the complex speech-to-motion mapping.
This issue becomes more pronounced with larger and more diverse datasets.
Additionally, many approaches \citep{faceformer2022, codetalker2023, multitalk2024} rely on a predefined set of style labels to reduce the complexity of the mapping, which limits their ability to adapt to new individuals and styles.

\subsection{Diffusion based 3D Facial Motion Generation}
Diffusion models \citep{ho2020denoising, zhu2023taming, alexanderson2023listen, facediffuser2023, aneja2023facetalk, baevski2020wav2vec, ma2024diffspeaker, diffposetalk2024,zhang2024letstalk} have recently gained traction for generative tasks because of their strong modeling capacities.
FaceDiffuser \citep{facediffuser2023} conditions a diffusion model on audio features to predict displacements from a neutral template.
Scantalk \citep{scantalk2024} uses a DiffusionNet \citep{sharp2022diffusionnet} structure to overcome the fixed topology limitation and perform diffusion on arbitrary meshes.
Similarly, DiffPoseTalk \citep{diffposetalk2024} uses a diffusion-based transformer decoder to generate both expression and pose (blendshapes) conditioned on audio and learned style features.
Although these methods generate high-fidelity and realistic facial motion, their iterative sampling steps can be computationally expensive and may limit real-time applicability.

In this paper, we leverage FLAME \citep{FLAME2017} as our motion representation and propose a novel autoregressive framework for speech-driven motion generation.
It not only outperform current diffusion models in generating facial motion, but also achieves real-time speed and produces natural head poses.
A comparison of our method with existing approaches is shown in Table \ref{tab:comp_methods}.
Our method enables arbitrary style inference, allowing stylized motion generation at inference time by providing an example motion clip.

\section{Method}
We provide an overview of our method in Figure \ref{fig:main_method}.
We adopt the widely used 3DMM \citep{FLAME2017} as the facial representation and leverage a multi-scale VQ autoencoder model to train and obtain a multi-scale motion codebook along with its corresponding encoder-decoder, enabling a discrete representation of the motion space.
Subsequently, we train a multi-scale autoregressive model to map speech information to the discrete motion space.

In the following subsections, Section \ref{sec:31} details the problem definition and the 3DMM representation, Section \ref{sec:32} explains the multi-scale motion VQ autoencoder model, Section \ref{sec:33} introduces the multi-scale autoregressive model.

\subsection{Preliminaries}
\label{sec:31}

We adopt the widely used 3D morphable model (3DMM) FLAME \citep{FLAME2017} to represent facial motion, where each frame is parameterized by shape \(\beta\), expression \(\psi\), and pose \(\theta\). Given these parameters, the 3D face mesh, consisting of 5,023 vertices, is reconstructed through blend shape and rotation operations. Compared to previous methods \citep{faceformer2022, codetalker2023, selftalk2023, scantalk2024} that directly model the mesh, FLAME simplifies motion modeling by converting complex vertex movements into lower-dimensional blend shape parameters while preserving expression and motion details.

We define the motion vector \(\mathbf{M} \in \mathbb{R}^{K \times D}\) as the concatenation of per-frame expression and pose parameters over \(K\) frames. The corresponding vertex coordinates \(\mathbf{V} \in \mathbb{R}^{K \times N \times 3}\) store the 3D positions of \(N\) mesh vertices per frame, where \(\mathbf{V}_{\mathrm{lips}} \subset \mathbf{V}\) denotes the subset corresponding to the lip region, which is particularly important for speech related modeling. Further details on 3DMM can be found in the supplementary material.

\subsection{Temporal Multi-scale VQ Autoencoder}
\label{sec:32}

Predicting motion frames from speech is a discrete task involving highly dense temporal sequences and complex mappings. Inspired by the success of discrete representations in image and motion generation \citep{codetalker2023, neuraldiscrete2017, zhou2022codeformer}, the effectiveness of multiscale audio modeling for animation generation \citep{jung2024speed} and the effectiveness of multi-scale encoding in \citep{var2024}, we propose a temporal multi-scale VQ autoencoder to efficiently model motion dynamics.

The input motion sequence, consisting of \(K\) frames, is first processed by a Transformer encoder, which extracts temporal features and maps them into a latent space. These representations are then quantized using a multi-scale codebook that captures motion information at varying temporal resolutions \(\bigl[k_1, k_2, \ldots, K\bigr]\), producing discrete tokens. The Transformer decoder reconstructs the motion from these discrete embeddings, attending to both local and long-range dependencies.

\noindent\textbf{Multi-scale residual VQ.}
To progressively refine motion representations, we adopt a residual vector quantization approach:
\begin{equation}
\label{eq:multi_scale_residual}
\begin{aligned}
    \mathbf{h}^{(l)} & = \mathrm{Interp}\bigl(\mathrm{Quant}(\mathbf{r}^{(l-1)}),\,k_l\bigr), \\
    \mathbf{r}^{(l)} & = \mathbf{r}^{(l-1)} - \mathbf{h}^{(l)},
\end{aligned}
\end{equation}
for \(l=1,\ldots,L\), where \(\mathbf{r}^{(0)}\) is the encoder output. The function \(\mathrm{Quant}(\cdot)\) assigns features to the closest codebook entries, while \(\mathrm{Interp}(\cdot,k_l)\) adjusts resolutions to ensure smooth transitions. By accumulating the outputs \(\mathbf{h}^{(l)}\) across scales, the model retains high fidelity while removing redundancy in a structured manner.

Residual accumulation at the finest scale \(K\) mitigates information loss, while downsampling via region-based interpolation enhances efficiency. We use a \emph{shared} codebook across scales to maintain a unified motion space and simplify autoregressive modeling. This strategy improves temporal consistency and preserves stylistic coherence across different resolutions.

\noindent\textbf{Temporal causal reasoning.}
To ensure stability over extended sequences, we incorporate a causal reasoning mechanism that enforces cross-window consistency. Our approach considers two consecutive time windows, \(T-1\) and \(T\). During training and inference, causal masks are applied, ensuring that predictions for \(T\) depend only on previously observed information from \(T-1\), without leaking future details. This prevents temporal discontinuities and enhances the smoothness of long-term motion.

By integrating multi-scale residual quantization with causal reasoning and Transformer-based encoding/decoding, our method ensures both fine-grained expressiveness and long-term coherence in generated motion sequences.

\noindent\textbf{Training objectives.}

To train the VQ autoencoder, we use a hybrid loss function balancing motion accuracy, temporal smoothness, and codebook stability.

The reconstruction loss enforces alignment between predicted motion vectors \(M'\) and the ground truth \(M\), with additional constraints on lip and facial vertices:
\begin{align}
\label{eq:vqloss1}
L_{\text{recon}} = \| \hat{M} - M \|_1 + w_{\text{lips}} \| \hat{V}_{\text{lips}} - V_{\text{lips}} \|^2 + \| \hat{V} - V \|^2
\end{align}
To ensure smooth transitions, we penalize velocity and acceleration differences:
\begin{align}
\label{eq:vqloss2}
L_{\text{vel}} &= \| (\hat{V}_{1:} - \hat{V}_{:-1}) - (V_{1:} - V_{:-1}) \|^2, \\
L_{\text{smooth}} &= \| \hat{V}_{2:} - 2\hat{V}_{1:-1} + \hat{V}_{:-2} \|^2.
\end{align}
We also apply the standard VQ losses which we term as \(L_{\text{cb}}\) to encourage stable VQ assignments and prevent mode collapse.

The final training objective is:
\begin{align}
\label{eq:vqloss_all}
L_{\text{VQ}} = L_{\text{recon}} + \lambda_{\text{vel}} L_{\text{vel}} + \lambda_{\text{smooth}} L_{\text{smooth}} + L_{\text{cb}},
\end{align}
where \(\lambda_{\text{vel}}\) and \(\lambda_{\text{smooth}}\) control the temporal constraints.

By jointly optimizing these terms, the model learns compact yet expressive motion representations, ensuring high-fidelity synthesis with strong temporal coherence.

\subsection{Speech-to-Motion Autoregressive Model}
\label{sec:33}

After training the VQ autoencoder (Section~\ref{sec:32}), we obtain discrete codes \(\{\mathbf{z}_{T}^{(l)}\}\) at multiple resolutions \(k_l\) for each time window \(T\), where \(l=1,\dots,L\) spans from coarsest to finest. To ensure long-term temporal coherence, we model these codes with a Transformer that is autoregressive \emph{across} scales \((1\!\to\!L)\) \emph{and across} adjacent windows \((T-1\!\to\!T)\).

\noindent\textbf{Architecture.}
Figure~\ref{fig:main_method}(b) illustrates the AR model, where orange arrows mark training-only components. A pre-trained HuBERT \citep{hsu2021hubert} encoder extracts speech features \(\mathbf{a}_T\), resampled to each scale \(k_l\). A style token \(\mathbf{s}\) encodes speaker identity \citep{diffposetalk2024}. Given the previous window’s finest-scale codes \(\mathbf{z}_{T-1}^{(L)}\), we predict \(\mathbf{z}_{T}^{(1)},\dots,\mathbf{z}_{T}^{(L)}\) for the current window.

\noindent\textbf{Two-level autoregression.}
Within each window \(T\), the codes are generated scale by scale:
\small
\begin{align}
&p(\{\mathbf{z}_{T}^{(l)}\}_{l=1}^L \mid \mathbf{z}_{T-1}^{(L)}, \mathbf{a}_T, \mathbf{s})
\;=\;
\prod_{l=1}^L 
p\!\bigl(\mathbf{z}_{T}^{(l)} \mid \mathbf{z}_{T}^{(<l)}, \mathbf{z}_{T-1}^{(L)}, \mathbf{a}_T, \mathbf{s}\bigr),
\end{align}
\normalsize
where \(\mathbf{z}_{T}^{(<l)}\) denotes codes at all coarser scales. Inside each scale \(l\), tokens are predicted \emph{in parallel} (blockwise) while attending causally to previously generated scales and the previous window.

\noindent\textbf{Training and inference.}
During training, ground-truth codes \(\{\mathbf{z}_{T}^{(l)\!\text{(gt)}}\}\) supervise the model via cross-entropy:
\begin{align}
L_{\mathrm{AR}}
&=\;
-\sum_{T}\sum_{l=1}^L 
\log\,
p(\mathbf{z}_{T}^{(l)\!\text{(gt)}} \mid \mathbf{z}_{T}^{(<l)\!\text{(gt)}}, \mathbf{z}_{T-1}^{(L)\!\text{(gt)}}, \mathbf{a}_T, \mathbf{s}).
\end{align}
At inference, we sample or select each scale’s codes in sequence, conditioned on \(\mathbf{z}_{T-1}^{(L)}\), \(\mathbf{a}_T\), and \(\mathbf{s}\). Repeating this process over windows yields motion sequences that remain temporally coherent and stylistically consistent throughout.

\begin{figure}[t]
\begin{center}
\centerline{\includegraphics[width=1.0\columnwidth]{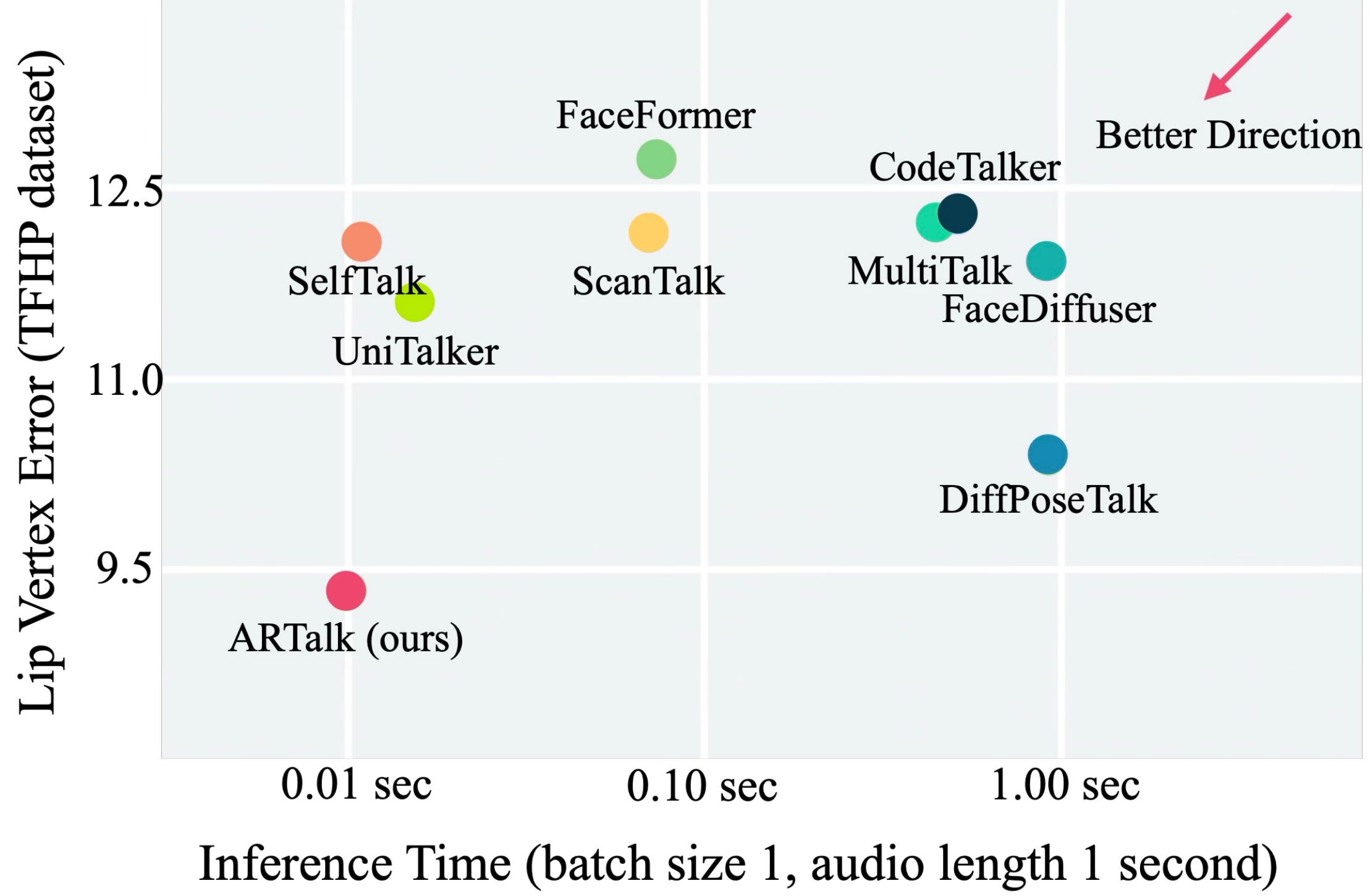}}
\caption{
Comparison of efficiency and performance across different methods.
}
\label{fig:comparison_speed}
\end{center}
\end{figure}

\section{Experiments}
\noindent\textbf{Datasets.}
We use the TFHP dataset proposed by ~\citet{diffposetalk2024} to train our model, which consists of 1,052 video clips from 588 subjects, with a total duration of approximately 26.5 hours.
All videos are tracked at 25 fps, resulting in approximately 2,385,000 action frames.
For training and evaluation, we adopt the train/test split provided in the original paper \citep{diffposetalk2024}.
Furthermore, we evaluate the generalization performance of our model on the test split of the widely used VOCASET dataset \citep{voca2019}.
The test split consists of 80 audiovisual sequence pairs from 2 subjects.
To adapt VOCASET to our method, we track the FLAME parameters at 25 fps and conduct evaluation on our tracked data.
It is worth noting that we don't use VOCASET to train or fine-tune our method.

\noindent\textbf{Implementation Details.}
Our method is implemented on the PyTorch framework \citep{pytorch2017}.
In the first stage, we train our VQ autoencoder to obtain a multi-scale motion codebook.
The motion codebook consists of 256 entries, each with a code dimension of 64.
The time window size is 100 frames (4 seconds), and the multi-scale levels are [1, 5, 25, 50, 100].
At this stage, we used the AdamW optimizer with a learning rate of 1.0e-4 and a total batch size of 64 for 50,000 iterations.
In the second stage, we train the multi-scale autoregressive model using the AdamW optimizer with the same learning rate of 1.0e-4 and a batch size of 64 for 50,000 iterations.
During this stage, we employ a frozen pre-trained HuBERT \citep{hsu2021hubert} backbone.

All training was conducted on one NVIDIA Tesla A100 GPU, requiring a total of approximately 13 GPU hours (5 hours for the first stage and 8 hours for the second stage), demonstrating efficient training resource utilization.
During inference, our method needs only 0.01 seconds to generate motions for 1 second on an NVIDIA Tesla A100 GPU and 0.057 seconds on an Apple M2 Pro chip, showcasing high inference efficiency and low latency.
For more implementation details, please refer to the supplementary materials.

\begin{table}[t]
\caption{
    Quantitative results on the TFHP \citep{diffposetalk2024} dataset. 
    We use colors to denote the \au{first} and \ag{second} places respectively.
    ${*}$ indicates that the method was not trained on TFHP, while all other methods were trained or fine-tuned on TFHP.
}
\label{tab:main_tfhp}
\tablestyle{7pt}{1.05}
\begin{center}
\scalebox{0.95}{
    \begin{tabular}{lccc}
    \toprule
    Method & LVE $\downarrow$ & FFD $\downarrow$ & MOD $\downarrow$ \\
    \midrule
    FaceFormer \citep{faceformer2022}       & 12.72 & 22.06 & 2.73  \\
    CodeTalker \citep{codetalker2023}       & 11.78 & 20.39 & 2.43  \\
    SelfTalk \citep{selftalk2023}           & 12.07 & 23.74 & 2.57  \\
    FaceDiffuser \citep{facediffuser2023}   & 11.92 & 22.17 & 2.55  \\
    MultiTalk$^{*}$ \citep{multitalk2024}   & 12.23 & 24.42 & 2.48  \\
    ScanTalk$^{*}$ \citep{scantalk2024}     & 12.14 & 21.02 & 3.20  \\
    UniTalker$^{*}$ \citep{scantalk2024}    & 11.61 & 29.31 & 2.07  \\
    DiffPoseTalk \citep{diffposetalk2024}   & \ag{10.39} & \ag{20.15} & \ag{2.07}  \\
    \midrule
    ARTalk (Ours)                           & \au{9.34} & \au{18.15} & \au{1.81} \\
    \bottomrule
    \end{tabular}
}
\end{center}
\end{table}

\begin{table}[t]
\caption{
    Quantitative results on the VOCASET-Test \citep{voca2019} dataset. 
    We use colors to denote the \au{first} and \ag{second} places respectively.
    ${*}$ indicates that the method was not trained on VOCASET.
    It is worth noting that our method is \textbf{not} trained or fine-tuned on VOCASET.
}
\label{tab:main_vocaset}
\tablestyle{7pt}{1.05}
\begin{center}
\scalebox{0.95}{
    \begin{tabular}{lccc}
    \toprule
    Method & LVE $\downarrow$ & FFD $\downarrow$ & MOD $\downarrow$ \\
    \midrule
    FaceFormer \citep{faceformer2022}           &  8.28 & \ag{15.62} & 1.89  \\
    CodeTalker \citep{codetalker2023}           &  7.73 & 18.86 & 1.81  \\
    SelfTalk \citep{selftalk2023}               &  7.71 & 28.70 & 1.79  \\
    FaceDiffuser \citep{facediffuser2023}       &  8.00 & 21.46 & 1.84  \\
    MultiTalk$^{*}$ \citep{multitalk2024}       & 12.33 & 25.11 & 2.82  \\
    ScanTalk \citep{scantalk2024}               &  \au{7.15} & 15.56 & \au{1.61}  \\
    UniTalker$^{*}$ \citep{scantalk2024}        &  7.38 & 21.72 & 1.89  \\
    DiffPoseTalk$^{*}$ \citep{diffposetalk2024} & 10.01 & 20.64 & 2.29  \\
    \midrule
    ARTalk (Ours)$^{*}$                           & \ag{7.57} & \au{15.49} & \ag{1.78} \\
    \bottomrule
    \end{tabular}
}
\end{center}
\end{table}

\begin{figure*}[ht]
\begin{center}
\centerline{\includegraphics[width=1.0\linewidth]{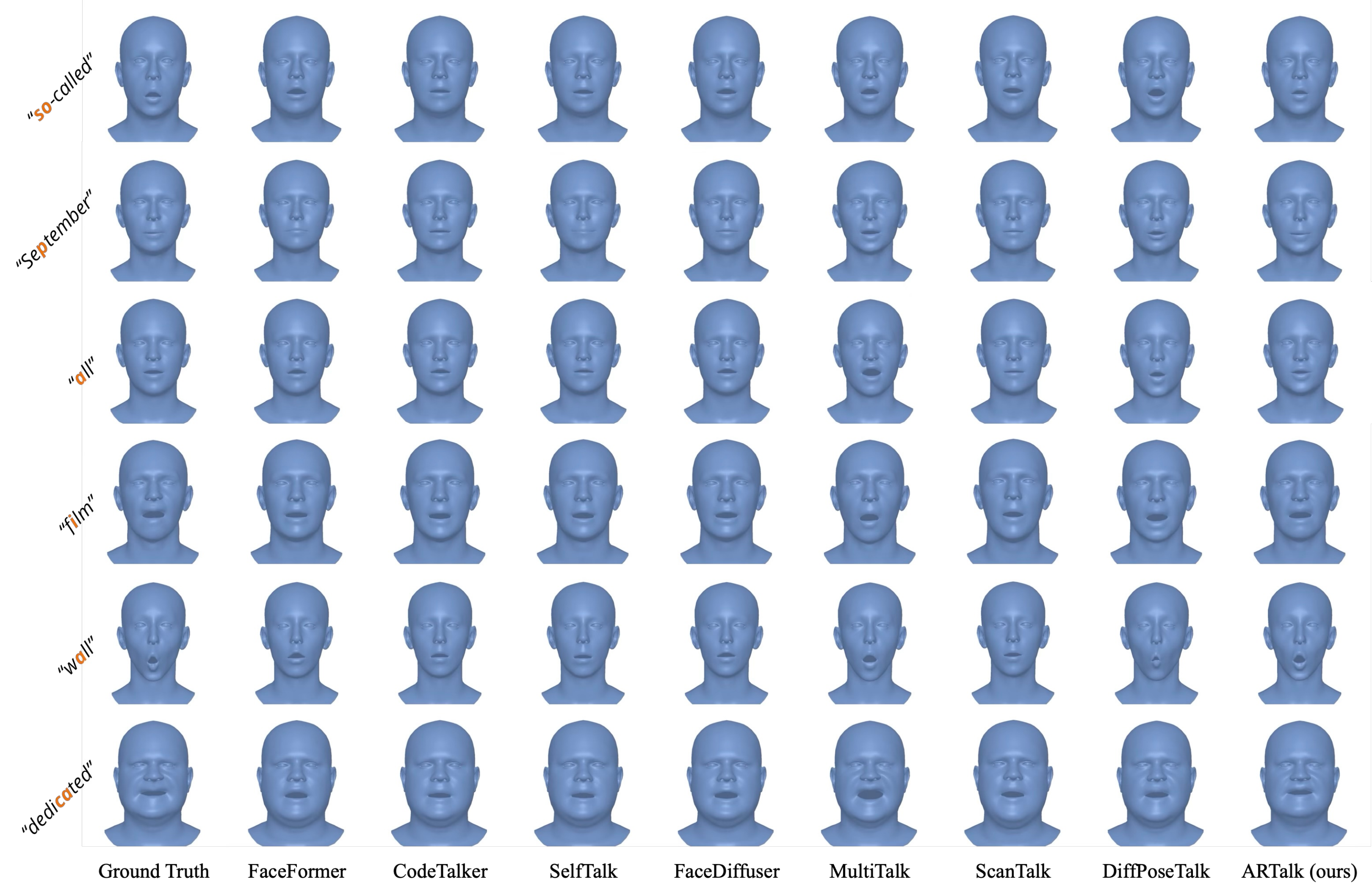}}
\caption{
Qualitative comparison with existing methods (all head poses fixed).
The first four rows are from the TFHP dataset, and the last two rows are from the VOCASET dataset.
Our method shows better alignment with the ground truth in expression style, mouth dynamics, and lip synchronization.
Additional results are available in the supplementary materials and demo videos.
}
\label{fig:main_results}
\end{center}
\Description{Qualitative comparison.}
\end{figure*}

\subsection{Quantitative Results}
Based on previous studies \citep{richard2021meshtalk, faceformer2022, codetalker2023, scantalk2024, diffposetalk2024}, we adopted two quantitative metrics, the lip vertex error (LVE) \citep{richard2021meshtalk} and the upper face dynamic deviation (FDD) \citep{codetalker2023}, to evaluate the generated facial motions.
LVE calculates the maximum L2 error of all lip vertices for each frame, evaluating the largest deviation between predicted and ground truth lip positions.
FDD calculates the standard deviation of the motion of each upper facial vertex over time between predictions and ground truth, evaluating the consistency of upper facial motion, which is closely related to speaking styles.
Additionally, we use a similar metric as ~\citet{diffposetalk2024}, mouth opening distance (MOD), to more accurately assess the stylistic similarity of mouth opening movements.
MOD measures the average difference in mouth opening region between predictions and ground truth.
Compared to LVE, MOD focuses more on the similarity of mouth opening styles and is less sensitive to temporal lip synchronization.
For the partitioning of the lip area and upper face, we used the mask partitioning officially provided by FLAME \citep{FLAME2017}.
The lip region contains 254 points, while the upper face, including eye region and forehead, contains 884 points.

We present the quantitative comparison on TFHP dataset \citep{diffposetalk2024} in Table \ref{tab:main_tfhp}. The baseline methods FaceFormer \citep{faceformer2022}, CodeTalker \citep{codetalker2023}, SelfTalk \citep{selftalk2023}, and FaceDiffuser \citep{facediffuser2023} are mesh-based approaches and cannot generalize to arbitrary styles or new meshes.
To ensure a fair comparison, we retrained these methods on meshes generated from the TFHP dataset.
For MultiTalk \citep{multitalk2024}, which is trained on FLAME meshes and supports language-based stylization, we used its English style for evaluation.
ScanTalk \citep{scantalk2024} claims to work on arbitrary meshes, so we use the officially provided pre-trained weights.
For UniTalker \citep{fan2024unitalker}, we compute the metrics using the meshes generated by the generalized pivot identities.

For DiffPoseTalk \citep{diffposetalk2024}, we used pre-trained weights on the TFHP dataset (including head pose) for evaluation.
The results in the Table \ref{tab:main_tfhp} show that our method achieving significant improvements in lip synchronization accuracy (LVE) and style alignment (FFD and MOD), indicating that our method not only achieves precise lip synchronization but also effectively captures personalized speaking styles.

To demonstrate the generalizability of our method, we evaluated it on the VOCASET dataset \citep{voca2019}.
The results are presented in Table \ref{tab:main_vocaset}.
Notably, for some baseline methods \citep{faceformer2022, codetalker2023, selftalk2023, facediffuser2023, scantalk2024}, we used their pre-trained weights on VOCASET, whereas our method was \textbf{not} trained or fine-tuned on this dataset.
Despite this, our method achieved highly competitive performance and outperformed most baseline methods specifically trained on this dataset.
For MultiTalk \citep{multitalk2024} and DiffPoseTalk \citep{diffposetalk2024} which designed with style generalization capabilities, we directly testing it on the VOCASET dataset without fine-tuning.
The results indicate that our method surpasses them in terms of generalization ability, further verifying the robustness of our method in handling unseen styles and data.

We also present a comparison of LVE and efficiency in Figure \ref{fig:comparison_speed}. 
Although our model adopts a two-level autoregressive framework, it remains more efficient than fully autoregressive methods.
This efficiency is achieved by utilizing longer window lengths while retaining only the past 4 seconds of motion frames.
Additionally, frames within each scale are generated in parallel within the window, further improving computational efficiency.

\subsection{Qualitative Results}
\begin{figure}[t]
\begin{center}
\centerline{\includegraphics[width=1.0\columnwidth]{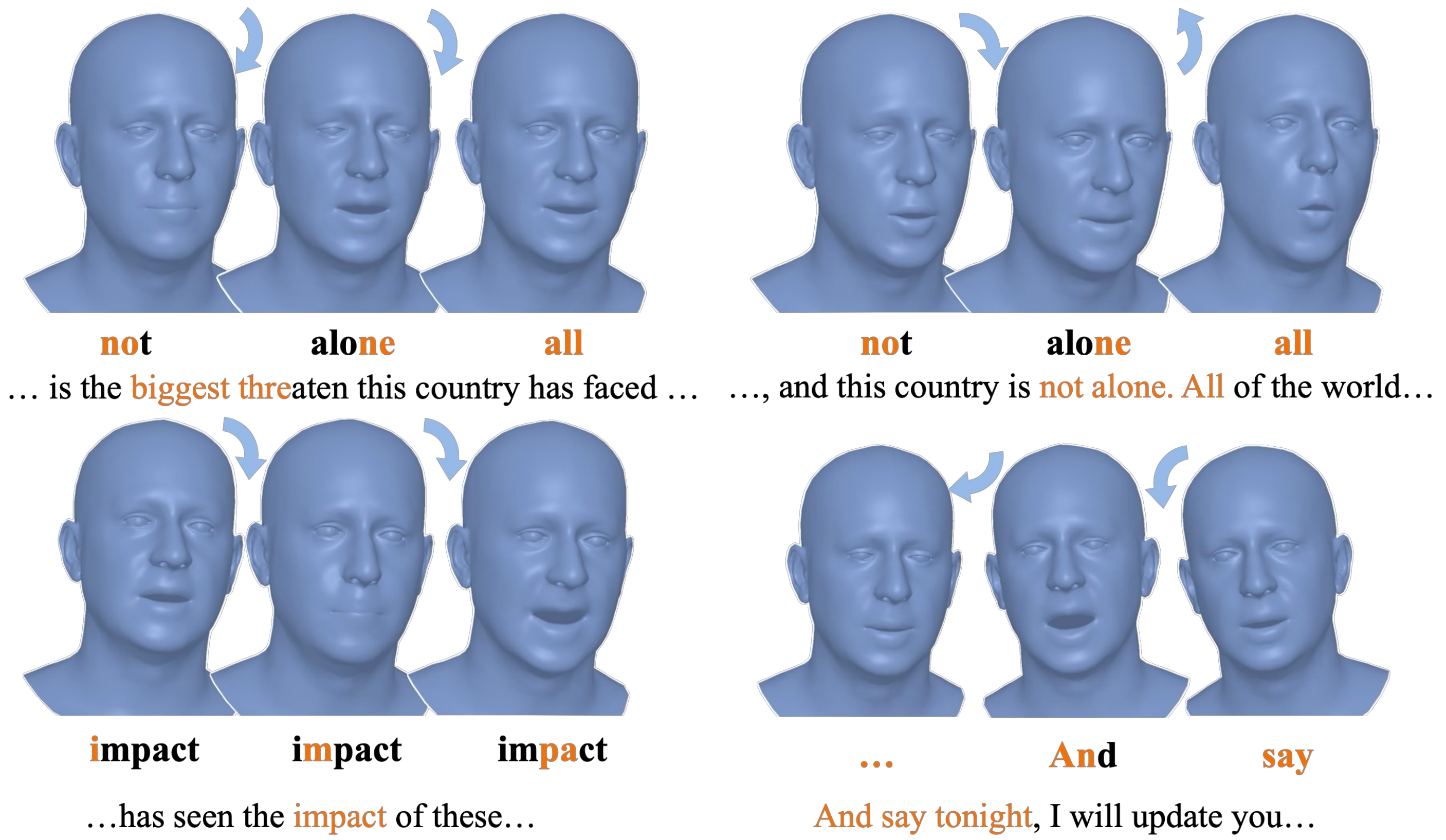}}
\caption{
Qualitative results of head pose.
When certain words are stressed or when accents occur, the model produces nodding motions similar to human behavior.
}
\label{fig:head_pose}
\end{center}
\end{figure}

In Figure \ref{fig:main_results}, we present a qualitative comparison between our method and other baseline approaches.
Our method demonstrates excellent lip synchronization, accurately capturing various phonetic elements.
Furthermore, the generated results exhibit realistic facial expressions and mouth opening, closely matching the style of the ground truth.
Notably, our method also generates realistic blinking and head movements, which are implicitly learned and encoded within the motion codebook.
In Figure \ref{fig:head_pose}, we also show the ability of our method on head movements. It can capture accents well and generate reasonable head movements.
Additional qualitative evaluation results are available in the supplementary videos.

\begin{table}[t]
\caption{
User study results.
The percentages represent \textbf{the proportion of users who preferred the our method over the baseline} in each category.
The values in brackets represent the 95\% confidence interval, indicating that we are 95\% confident that the true value of the performance metric lies within this range.
Among them, Sync represents lip synchronization, Style represents the consistency of style, N-Exp and N-Pose represent the facial expression naturalness and head pose naturalness respectively.
}
\label{tab:user_study}
\tablestyle{3pt}{1.05}
\begin{center}
\scalebox{0.79}{
    \begin{tabular}{lcccc}
    \toprule
    Method & Sync (\%) & N-Exp (\%) & Style (\%) & N-Pose (\%) \\
    \midrule
    FaceFormer      & 75.0 [70.4, 79.6] & 89.3 [86.3, 92.8] & 88.1 [85.0, 91.8] & - \\
    CodeTalker      & 84.5 [80.7, 88.4] & 86.9 [83.3, 90.5] & 92.9 [90.5, 95.9] & - \\
    SelfTalk        & 85.7 [82.0, 89.5] & 89.3 [86.3, 92.8] & 91.7 [89.1, 94.9] & - \\
    FaceDiffuser    & 88.1 [85.0, 91.8] & 90.4 [87.3, 93.6] & 86.9 [83.3, 90.5] & - \\
    MultiTalk       & 78.6 [74.5, 83.2] & 76.2 [72.0, 81.0] & 78.6 [74.5, 83.2] & - \\
    ScanTalk        & 88.1 [85.0, 91.8] & 90.5 [87.7, 93.9] & 90.5 [87.7, 93.9] & - \\
    UniTalker       & 81.8 [77.7, 86.0] & 83.6 [79.7, 87.6] & 81.3 [77.4, 85.7] & - \\
    DiffPoseTalk    & 63.1 [58.2, 68.5] & 59.5 [54.3, 64.8] & 60.7 [55.5, 65.9] & 58.3 [53.1, 63.6] \\
    \bottomrule
    \end{tabular}
}
\end{center}
\end{table}

\subsection{User Study}
User studies are a reliable approach for evaluating 3D talking heads.
To comprehensively compare our method with baseline methods, we conducted a user study focusing on four key metrics: lip synchronization, facial expression naturalness, style consistency and head pose naturalness.

Given that head movements can significantly influence users' evaluation of lip synchronization, the study was divided into two parts.
In the first part, videos with fixed head poses were rendered across all methods to assess lip synchronization and facial expression realism and style consistency.
In the second part, videos with dynamic head poses were rendered to evaluate style consistency and head pose naturalness.
All comparisons were conducted using pairwise comparisons, where motions generated by our method and a competing baseline were displayed side by side, along with the ground truth provided as a reference for users.
After watching the videos, users selected the animation they perceived as better based on their subjective preference.
The proportion of user selections was calculated to quantify satisfaction.

As presented in Table \ref{tab:user_study}, our method significantly outperformed the baseline methods in lip synchronization, style consistency and facial expression realism.
Furthermore, our method achieved superior head pose naturalness compared to DiffPoseTalk \citep{diffposetalk2024}, demonstrating its comprehensive advantages.

\subsection{Ablation study}
\noindent\textbf{Multi-Scale Autoregression.}
To validate the importance of multi-scale autoregressive modeling within a single time window, we conducted experiments by removing the multi-scale autoregressive mechanism entirely. The results shown in Table \ref{tab:ablation} indicate a significant drop in generation accuracy without multi-scale autoregression.
We attribute this decline to the model's inability to fully capture the complex details of the speech-to-motion mapping without the hierarchical structure provided by multi-scale modeling.

\noindent\textbf{Temporal Encoder and Temporal Autoregression.}
To assess the necessity of temporal encoding and autoregression across time windows, we replaced our proposed temporal encoder and autoregressive model with a standard multi-scale encoder and a single-window multi-scale autoregressive model, respectively.
As shown in Table \ref{tab:ablation}, the absence of temporal modeling in VQ autoencoder or autoregressive both result in degraded generation quality.
Additionally, we also observed temporal discontinuities in the motion sequences, further emphasizing the critical role of our temporal design in achieving smooth and coherent outputs.

\noindent\textbf{Speaker Style Embedding.}
We also evaluated the effect of removing the speaker style feature, which serves as the starting condition for the autoregressive process.
Without this feature, the complexity of the many-to-many mapping between speech and motion increases significantly, and the generated motions lack personalized style.
As reflected in Table \ref{tab:ablation}, the removal of style features causes a substantial decline in generation quality, highlighting their importance in achieving stylistic and expressive outputs.

\noindent\textbf{Key Hyperparameter Choices.}
We explored different number of motion frames within a single time window, which is critical for our method.
Specifically, we tested window lengths of 8, 25, 50 frames.
For a window length of 8, we used a multi-scale sequence of [1,2,4,8], while for 25 and 50, we adopted [1,5,10,15,25] and [1,5,10,25,50], respectively.
The results, presented in Table \ref{tab:ablation}, demonstrate that the choice of window size significantly impacts performance.
However, the window size depends heavily on the downstream task characteristics, such as whether the input consists of streaming audio chunks.
In general, longer temporal windows generally produce better results, but longer windows mean that more audio needs to be collected to output motion.
In this paper, we chose a window length of 100 frames and a multi-scale sequence of [1, 5, 25, 50, 100] as it is consistent with previous work such as DiffPoseTalk \citep{diffposetalk2024} and achieves a good balance between quality and efficiency.
Further exploration of the number of multi-scale layers and audio chunk lengths is provided in the supplementary material.

\begin{figure}
\begin{center}
\centerline{\includegraphics[width=1.0\linewidth]{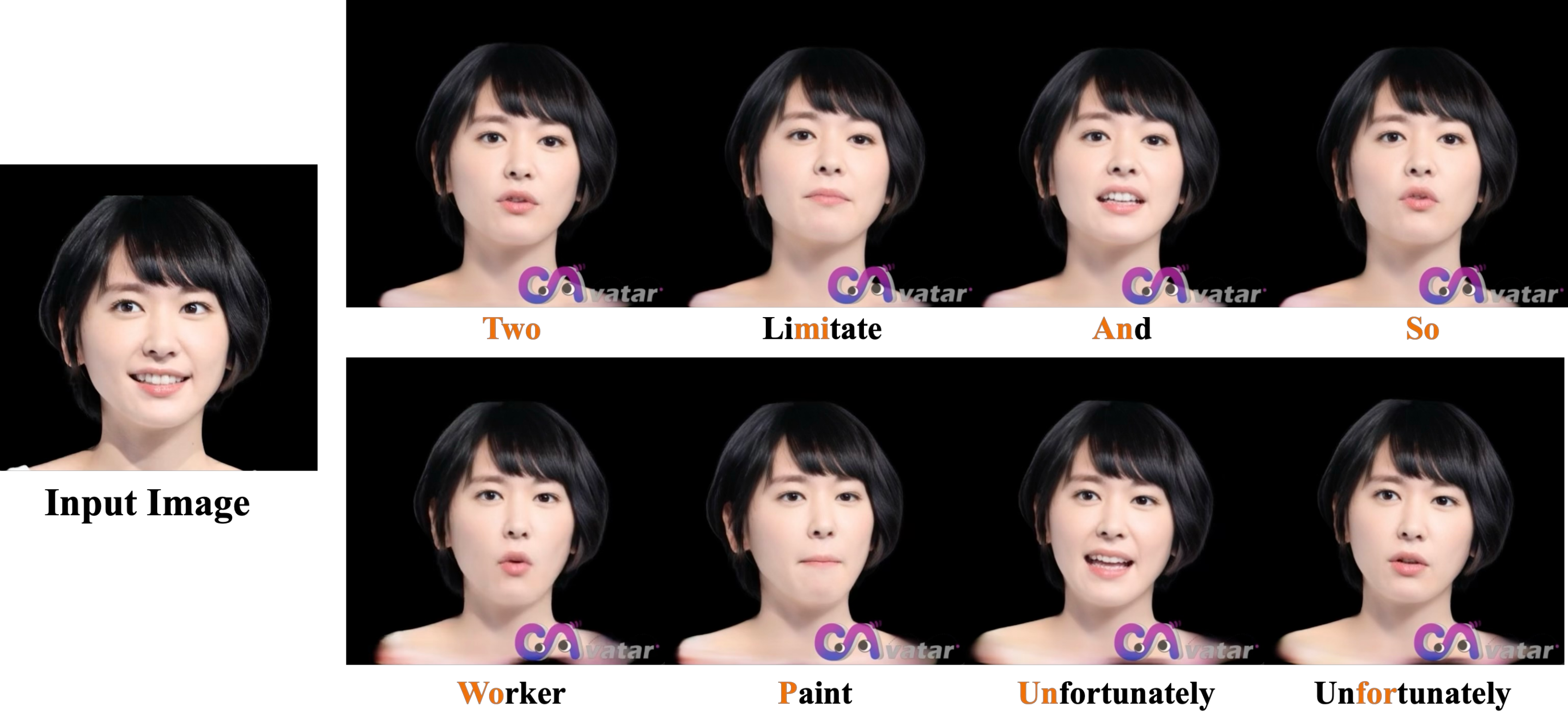}}
\caption{
We reconstruct the avatar using GAGAvatar \citep{chu2024gagavatar} and drive it with our ARTalk model, enabling speech-driven dynamic avatar generation.
}
\label{fig:supp_gaga}
\end{center}
\end{figure}

\noindent\textbf{Integration with Downstream Tasks}
Our method can be integrated into various FLAME-based downstream applications, enabling a wider range of use cases. 
While this extends beyond the primary scope of this paper, we also demonstrate its application in GAGAvatar \citep{chu2024gagavatar}.
As shown in Figure \ref{fig:supp_gaga}, our method can generate control signals for GAGAvatar, effectively transforming it into a speech-driven one-shot dynamic avatar reconstruction method.

\begin{table}[t]
\caption{Ablation results on TFHP dataset.}
\label{tab:ablation}
\tablestyle{12pt}{1.05}
\begin{center}
\scalebox{0.95}{
    \begin{tabular}{lccc}
    \toprule
    Method & LVE$\downarrow$ & FFD$\downarrow$ & MOD $\downarrow$ \\
    \midrule
    w/o Multi-Scale AR      & 14.14 & 22.04 & 3.06 \\
    w/o Temporal VQ         &  9.82 & 18.60 & 1.86 \\
    w/o Temporal AR         &  9.97 & 18.71 & 1.99 \\
    w/o Style Embedding     & 11.80 & 21.46 & 2.37 \\
    Clip Length 8 Layer 5    & 11.73 & 24.89 & 2.25 \\ 
    Clip Length 25 Layer 5   & 10.20 & 19.00 & 1.97 \\ 
    Clip Length 50 Layer 5   &  9.78 & \textbf{18.03} & 1.89 \\ 
    \midrule
    ARTalk (Full)           &  9.34 & 18.15 & \textbf{1.81} \\
    \bottomrule
    \end{tabular}
}
\end{center}
\end{table}

\section{Discussion and Conclusion}
In this paper, we introduced ARTalk, a novel framework for generating 3D facial and head motions from speech.  
The core innovation of our method lies in the temporal multi-scale autoencoder and the ARTalk autoregressive Transformer, which together ensure temporal consistency and precise motion generation.
Our experimental results demonstrate that ARTalk outperforms state-of-the-art baselines in terms of lip synchronization, expression naturalness and style consistency, while maintaining real-time generation capabilities.  
We believe that the strong generalization ability of ARTalk make it a promising solution for a wide range of applications, including virtual avatars, language training, and animation production for gaming and movies.

\noindent\textbf{Limitations and future work.}
While ARTalk demonstrates strong performance in lip synchronization and expressions, a key limitation is that head movements are driven by speech prosody, not semantic context.
This prevents the model from generating culturally-specific or meaningful gestures like nodding or shaking the head to indicate affirmation.
Additionally, our use of a single style embedding can be challenging for speech with dramatic emotional shifts, which is also a common problem with existing methods.
For example, when a speaker suddenly shifts from calm to excited, a single embedding may struggle to accurately capture this transition.

We hope that our work lays a solid foundation for the further research of 3D talking head generation.
Future research will focus on addressing the above challenges, especially exploring how to effectively incorporate semantic information into head motion generation and developing more flexible style control methods.

\section*{Impact Statement}
Given that our method can generate highly realistic talking head sequences, there is a potential risk of misuse, such as deepfake creation and manipulation.
To address these concerns, we strongly advocate for watermarking or clearly annotating facial sequences generated by our method as synthetic data.  
Moreover, we will collaborate on deepfake detection research to further ensure that our proposed method is used for positive and constructive applications.
We also aim to raise awareness about these risks and encourage collaboration among governments, developers, researchers, and users to jointly prevent the misuse of synthetic video technologies.

\begin{acks}
This work was partially supported by JST Moonshot R\&D Grant Number JPMJPS2011, CREST Grant Number JPMJCR2015 and Basic Research Grant (Super AI) of Institute for AI and Beyond of the University of Tokyo.
In addition, this work was also partially supported by JST SPRING, Grant Number JPMJSP2108.
\end{acks}

\bibliographystyle{ACM-Reference-Format}
\bibliography{bibliography}


\begin{thebibliography}{54}


\ifx \showCODEN    \undefined \def \showCODEN     #1{\unskip}     \fi
\ifx \showDOI      \undefined \def \showDOI       #1{#1}\fi
\ifx \showISBNx    \undefined \def \showISBNx     #1{\unskip}     \fi
\ifx \showISBNxiii \undefined \def \showISBNxiii  #1{\unskip}     \fi
\ifx \showISSN     \undefined \def \showISSN      #1{\unskip}     \fi
\ifx \showLCCN     \undefined \def \showLCCN      #1{\unskip}     \fi
\ifx \shownote     \undefined \def \shownote      #1{#1}          \fi
\ifx \showarticletitle \undefined \def \showarticletitle #1{#1}   \fi
\ifx \showURL      \undefined \def \showURL       {\relax}        \fi
\providecommand\bibfield[2]{#2}
\providecommand\bibinfo[2]{#2}
\providecommand\natexlab[1]{#1}
\providecommand\showeprint[2][]{arXiv:#2}

\bibitem[Achiam et~al\mbox{.}(2023)]%
        {achiam2023gpt}
\bibfield{author}{\bibinfo{person}{Josh Achiam}, \bibinfo{person}{Steven Adler}, \bibinfo{person}{Sandhini Agarwal}, \bibinfo{person}{Lama Ahmad}, \bibinfo{person}{Ilge Akkaya}, \bibinfo{person}{Florencia~Leoni Aleman}, \bibinfo{person}{Diogo Almeida}, \bibinfo{person}{Janko Altenschmidt}, \bibinfo{person}{Sam Altman}, \bibinfo{person}{Shyamal Anadkat}, {et~al\mbox{.}}} \bibinfo{year}{2023}\natexlab{}.
\newblock \showarticletitle{Gpt-4 technical report}.
\newblock \bibinfo{journal}{\emph{arXiv preprint arXiv:2303.08774}} (\bibinfo{year}{2023}).
\newblock


\bibitem[Alexanderson et~al\mbox{.}(2023)]%
        {alexanderson2023listen}
\bibfield{author}{\bibinfo{person}{Simon Alexanderson}, \bibinfo{person}{Rajmund Nagy}, \bibinfo{person}{Jonas Beskow}, {and} \bibinfo{person}{Gustav~Eje Henter}.} \bibinfo{year}{2023}\natexlab{}.
\newblock \showarticletitle{Listen, denoise, action! audio-driven motion synthesis with diffusion models}.
\newblock \bibinfo{journal}{\emph{ACM Transactions on Graphics (TOG)}} \bibinfo{volume}{42}, \bibinfo{number}{4} (\bibinfo{year}{2023}), \bibinfo{pages}{1--20}.
\newblock


\bibitem[Aneja et~al\mbox{.}(2024)]%
        {aneja2023facetalk}
\bibfield{author}{\bibinfo{person}{Shivangi Aneja}, \bibinfo{person}{Justus Thies}, \bibinfo{person}{Angela Dai}, {and} \bibinfo{person}{Matthias Nießner}.} \bibinfo{year}{2024}\natexlab{}.
\newblock \showarticletitle{FaceTalk: Audio-Driven Motion Diffusion for Neural Parametric Head Models}. In \bibinfo{booktitle}{\emph{Proc. IEEE Conf. on Computer Vision and Pattern Recognition (CVPR)}}.
\newblock


\bibitem[Baevski et~al\mbox{.}(2020)]%
        {baevski2020wav2vec}
\bibfield{author}{\bibinfo{person}{Alexei Baevski}, \bibinfo{person}{Yuhao Zhou}, \bibinfo{person}{Abdelrahman Mohamed}, {and} \bibinfo{person}{Michael Auli}.} \bibinfo{year}{2020}\natexlab{}.
\newblock \showarticletitle{wav2vec 2.0: A framework for self-supervised learning of speech representations}.
\newblock \bibinfo{journal}{\emph{Advances in neural information processing systems}}  \bibinfo{volume}{33} (\bibinfo{year}{2020}), \bibinfo{pages}{12449--12460}.
\newblock


\bibitem[Chae-Yeon et~al\mbox{.}(2025)]%
        {chae2025perceptually}
\bibfield{author}{\bibinfo{person}{Lee Chae-Yeon}, \bibinfo{person}{Oh Hyun-Bin}, \bibinfo{person}{Han EunGi}, \bibinfo{person}{Kim Sung-Bin}, \bibinfo{person}{Suekyeong Nam}, {and} \bibinfo{person}{Tae-Hyun Oh}.} \bibinfo{year}{2025}\natexlab{}.
\newblock \showarticletitle{Perceptually Accurate 3D Talking Head Generation: New Definitions, Speech-Mesh Representation, and Evaluation Metrics}.
\newblock \bibinfo{journal}{\emph{arXiv preprint arXiv:2503.20308}} (\bibinfo{year}{2025}).
\newblock


\bibitem[Chu and Harada(2024)]%
        {chu2024gagavatar}
\bibfield{author}{\bibinfo{person}{Xuangeng Chu} {and} \bibinfo{person}{Tatsuya Harada}.} \bibinfo{year}{2024}\natexlab{}.
\newblock \showarticletitle{Generalizable and Animatable Gaussian Head Avatar}. In \bibinfo{booktitle}{\emph{The Thirty-eighth Annual Conference on Neural Information Processing Systems}}.
\newblock
\urldef\tempurl%
\url{https://openreview.net/forum?id=gVM2AZ5xA6}
\showURL{%
\tempurl}


\bibitem[Chu et~al\mbox{.}(2024)]%
        {gpavatar2024}
\bibfield{author}{\bibinfo{person}{Xuangeng Chu}, \bibinfo{person}{Yu Li}, \bibinfo{person}{Ailing Zeng}, \bibinfo{person}{Tianyu Yang}, \bibinfo{person}{Lijian Lin}, \bibinfo{person}{Yunfei Liu}, {and} \bibinfo{person}{Tatsuya Harada}.} \bibinfo{year}{2024}\natexlab{}.
\newblock \showarticletitle{{GPA}vatar: Generalizable and Precise Head Avatar from Image(s)}. In \bibinfo{booktitle}{\emph{The Twelfth International Conference on Learning Representations}}.
\newblock


\bibitem[Cudeiro et~al\mbox{.}(2019)]%
        {voca2019}
\bibfield{author}{\bibinfo{person}{Daniel Cudeiro}, \bibinfo{person}{Timo Bolkart}, \bibinfo{person}{Cassidy Laidlaw}, \bibinfo{person}{Anurag Ranjan}, {and} \bibinfo{person}{Michael Black}.} \bibinfo{year}{2019}\natexlab{}.
\newblock \showarticletitle{Capture, Learning, and Synthesis of {3D} Speaking Styles}. In \bibinfo{booktitle}{\emph{Proceedings IEEE Conf. on Computer Vision and Pattern Recognition (CVPR)}}. \bibinfo{pages}{10101--10111}.
\newblock
\urldef\tempurl%
\url{http://voca.is.tue.mpg.de/}
\showURL{%
\tempurl}


\bibitem[Danecek et~al\mbox{.}(2022)]%
        {EMOCA2021}
\bibfield{author}{\bibinfo{person}{Radek Danecek}, \bibinfo{person}{Michael~J. Black}, {and} \bibinfo{person}{Timo Bolkart}.} \bibinfo{year}{2022}\natexlab{}.
\newblock \showarticletitle{{EMOCA}: {E}motion Driven Monocular Face Capture and Animation}. In \bibinfo{booktitle}{\emph{Conference on Computer Vision and Pattern Recognition (CVPR)}}. \bibinfo{pages}{20311--20322}.
\newblock


\bibitem[Dan{\v{e}}{\v{c}}ek et~al\mbox{.}(2023)]%
        {danvevcek2023emotional}
\bibfield{author}{\bibinfo{person}{Radek Dan{\v{e}}{\v{c}}ek}, \bibinfo{person}{Kiran Chhatre}, \bibinfo{person}{Shashank Tripathi}, \bibinfo{person}{Yandong Wen}, \bibinfo{person}{Michael Black}, {and} \bibinfo{person}{Timo Bolkart}.} \bibinfo{year}{2023}\natexlab{}.
\newblock \showarticletitle{Emotional speech-driven animation with content-emotion disentanglement}. In \bibinfo{booktitle}{\emph{SIGGRAPH Asia 2023 Conference Papers}}. \bibinfo{pages}{1--13}.
\newblock


\bibitem[D\'efossez et~al\mbox{.}(2024)]%
        {kyutai2024moshi}
\bibfield{author}{\bibinfo{person}{Alexandre D\'efossez}, \bibinfo{person}{Laurent Mazar\'e}, \bibinfo{person}{Manu Orsini}, \bibinfo{person}{Am\'elie Royer}, \bibinfo{person}{Patrick P\'erez}, \bibinfo{person}{Herv\'e J\'egou}, \bibinfo{person}{Edouard Grave}, {and} \bibinfo{person}{Neil Zeghidour}.} \bibinfo{year}{2024}\natexlab{}.
\newblock \bibinfo{booktitle}{\emph{Moshi: a speech-text foundation model for real-time dialogue}}.
\newblock \bibinfo{type}{{T}echnical {R}eport}. \bibinfo{institution}{Kyutai}.
\newblock
\urldef\tempurl%
\url{http://kyutai.org/Moshi.pdf}
\showURL{%
\tempurl}


\bibitem[Deng et~al\mbox{.}(2024)]%
        {deng2024portrait4d2}
\bibfield{author}{\bibinfo{person}{Yu Deng}, \bibinfo{person}{Duomin Wang}, {and} \bibinfo{person}{Baoyuan Wang}.} \bibinfo{year}{2024}\natexlab{}.
\newblock \showarticletitle{Portrait4D-v2: Pseudo Multi-View Data Creates Better 4D Head Synthesizer}.
\newblock \bibinfo{journal}{\emph{arXiv preprint arXiv:2403.13570}} (\bibinfo{year}{2024}).
\newblock


\bibitem[Edwards et~al\mbox{.}(2016)]%
        {edwards2016jali}
\bibfield{author}{\bibinfo{person}{Pif Edwards}, \bibinfo{person}{Chris Landreth}, \bibinfo{person}{Eugene Fiume}, {and} \bibinfo{person}{Karan Singh}.} \bibinfo{year}{2016}\natexlab{}.
\newblock \showarticletitle{Jali: an animator-centric viseme model for expressive lip synchronization}.
\newblock \bibinfo{journal}{\emph{ACM Transactions on graphics (TOG)}} \bibinfo{volume}{35}, \bibinfo{number}{4} (\bibinfo{year}{2016}), \bibinfo{pages}{1--11}.
\newblock


\bibitem[Fan et~al\mbox{.}(2024)]%
        {fan2024unitalker}
\bibfield{author}{\bibinfo{person}{Xiangyu Fan}, \bibinfo{person}{Jiaqi Li}, \bibinfo{person}{Zhiqian Lin}, \bibinfo{person}{Weiye Xiao}, {and} \bibinfo{person}{Lei Yang}.} \bibinfo{year}{2024}\natexlab{}.
\newblock \showarticletitle{UniTalker: Scaling up Audio-Driven 3D Facial Animation through A Unified Model}. In \bibinfo{booktitle}{\emph{European Conference on Computer Vision}}. Springer, \bibinfo{pages}{204--221}.
\newblock


\bibitem[Fan et~al\mbox{.}(2022)]%
        {faceformer2022}
\bibfield{author}{\bibinfo{person}{Yingruo Fan}, \bibinfo{person}{Zhaojiang Lin}, \bibinfo{person}{Jun Saito}, \bibinfo{person}{Wenping Wang}, {and} \bibinfo{person}{Taku Komura}.} \bibinfo{year}{2022}\natexlab{}.
\newblock \showarticletitle{FaceFormer: Speech-Driven 3D Facial Animation with Transformers}. In \bibinfo{booktitle}{\emph{Proceedings of the IEEE/CVF Conference on Computer Vision and Pattern Recognition (CVPR)}}.
\newblock


\bibitem[Feng et~al\mbox{.}(2021)]%
        {DECA2021}
\bibfield{author}{\bibinfo{person}{Yao Feng}, \bibinfo{person}{Haiwen Feng}, \bibinfo{person}{Michael~J. Black}, {and} \bibinfo{person}{Timo Bolkart}.} \bibinfo{year}{2021}\natexlab{}.
\newblock \showarticletitle{Learning an Animatable Detailed {3D} Face Model from In-The-Wild Images}.
\newblock \bibinfo{journal}{\emph{ACM Transactions on Graphics, (Proc. SIGGRAPH)}} \bibinfo{volume}{40}, \bibinfo{number}{8}.
\newblock
\urldef\tempurl%
\url{https://doi.org/10.1145/3450626.3459936}
\showURL{%
\tempurl}


\bibitem[Guo et~al\mbox{.}(2025)]%
        {guo2025deepseek}
\bibfield{author}{\bibinfo{person}{Daya Guo}, \bibinfo{person}{Dejian Yang}, \bibinfo{person}{Haowei Zhang}, \bibinfo{person}{Junxiao Song}, \bibinfo{person}{Ruoyu Zhang}, \bibinfo{person}{Runxin Xu}, \bibinfo{person}{Qihao Zhu}, \bibinfo{person}{Shirong Ma}, \bibinfo{person}{Peiyi Wang}, \bibinfo{person}{Xiao Bi}, {et~al\mbox{.}}} \bibinfo{year}{2025}\natexlab{}.
\newblock \showarticletitle{Deepseek-r1: Incentivizing reasoning capability in llms via reinforcement learning}.
\newblock \bibinfo{journal}{\emph{arXiv preprint arXiv:2501.12948}} (\bibinfo{year}{2025}).
\newblock


\bibitem[Ho et~al\mbox{.}(2020)]%
        {ho2020denoising}
\bibfield{author}{\bibinfo{person}{Jonathan Ho}, \bibinfo{person}{Ajay Jain}, {and} \bibinfo{person}{Pieter Abbeel}.} \bibinfo{year}{2020}\natexlab{}.
\newblock \showarticletitle{Denoising diffusion probabilistic models}.
\newblock \bibinfo{journal}{\emph{Advances in neural information processing systems}}  \bibinfo{volume}{33} (\bibinfo{year}{2020}), \bibinfo{pages}{6840--6851}.
\newblock


\bibitem[Hsu et~al\mbox{.}(2021)]%
        {hsu2021hubert}
\bibfield{author}{\bibinfo{person}{Wei-Ning Hsu}, \bibinfo{person}{Benjamin Bolte}, \bibinfo{person}{Yao-Hung~Hubert Tsai}, \bibinfo{person}{Kushal Lakhotia}, \bibinfo{person}{Ruslan Salakhutdinov}, {and} \bibinfo{person}{Abdelrahman Mohamed}.} \bibinfo{year}{2021}\natexlab{}.
\newblock \showarticletitle{Hubert: Self-supervised speech representation learning by masked prediction of hidden units}.
\newblock \bibinfo{journal}{\emph{IEEE/ACM transactions on audio, speech, and language processing}}  \bibinfo{volume}{29} (\bibinfo{year}{2021}), \bibinfo{pages}{3451--3460}.
\newblock


\bibitem[Huang and Belongie(2017)]%
        {huang2017arbitrary}
\bibfield{author}{\bibinfo{person}{Xun Huang} {and} \bibinfo{person}{Serge Belongie}.} \bibinfo{year}{2017}\natexlab{}.
\newblock \showarticletitle{Arbitrary style transfer in real-time with adaptive instance normalization}. In \bibinfo{booktitle}{\emph{Proceedings of the IEEE international conference on computer vision}}. \bibinfo{pages}{1501--1510}.
\newblock


\bibitem[Jung et~al\mbox{.}(2024)]%
        {jung2024speed}
\bibfield{author}{\bibinfo{person}{Sunjin Jung}, \bibinfo{person}{Yeongho Seol}, \bibinfo{person}{Kwanggyoon Seo}, \bibinfo{person}{Hyeonho Na}, \bibinfo{person}{Seonghyeon Kim}, \bibinfo{person}{Vanessa Tan}, {and} \bibinfo{person}{Junyong Noh}.} \bibinfo{year}{2024}\natexlab{}.
\newblock \showarticletitle{Speed-aware audio-driven speech animation using adaptive windows}.
\newblock \bibinfo{journal}{\emph{ACM Transactions on Graphics}} \bibinfo{volume}{44}, \bibinfo{number}{1} (\bibinfo{year}{2024}), \bibinfo{pages}{1--14}.
\newblock


\bibitem[Li et~al\mbox{.}(2017)]%
        {FLAME2017}
\bibfield{author}{\bibinfo{person}{Tianye Li}, \bibinfo{person}{Timo Bolkart}, \bibinfo{person}{Michael.~J. Black}, \bibinfo{person}{Hao Li}, {and} \bibinfo{person}{Javier Romero}.} \bibinfo{year}{2017}\natexlab{}.
\newblock \showarticletitle{Learning a model of facial shape and expression from {4D} scans}.
\newblock \bibinfo{journal}{\emph{ACM Transactions on Graphics, (Proc. SIGGRAPH Asia)}} \bibinfo{volume}{36}, \bibinfo{number}{6} (\bibinfo{year}{2017}), \bibinfo{pages}{194:1--194:17}.
\newblock
\urldef\tempurl%
\url{https://doi.org/10.1145/3130800.3130813}
\showURL{%
\tempurl}


\bibitem[Loshchilov(2017)]%
        {loshchilov2017decoupled}
\bibfield{author}{\bibinfo{person}{I Loshchilov}.} \bibinfo{year}{2017}\natexlab{}.
\newblock \showarticletitle{Decoupled weight decay regularization}.
\newblock \bibinfo{journal}{\emph{arXiv preprint arXiv:1711.05101}} (\bibinfo{year}{2017}).
\newblock


\bibitem[Ma et~al\mbox{.}(2024)]%
        {ma2024diffspeaker}
\bibfield{author}{\bibinfo{person}{Zhiyuan Ma}, \bibinfo{person}{Xiangyu Zhu}, \bibinfo{person}{Guojun Qi}, \bibinfo{person}{Chen Qian}, \bibinfo{person}{Zhaoxiang Zhang}, {and} \bibinfo{person}{Zhen Lei}.} \bibinfo{year}{2024}\natexlab{}.
\newblock \showarticletitle{DiffSpeaker: Speech-Driven 3D Facial Animation with Diffusion Transformer}.
\newblock \bibinfo{journal}{\emph{CoRR}}  \bibinfo{volume}{abs/2402.05712} (\bibinfo{year}{2024}).
\newblock
\urldef\tempurl%
\url{https://doi.org/10.48550/arXiv.2402.05712}
\showURL{%
\tempurl}


\bibitem[Nocentini et~al\mbox{.}(2024)]%
        {scantalk2024}
\bibfield{author}{\bibinfo{person}{F. Nocentini}, \bibinfo{person}{T. Besnier}, \bibinfo{person}{C. Ferrari}, \bibinfo{person}{S. Arguillere}, \bibinfo{person}{S. Berretti}, {and} \bibinfo{person}{M. Daoudi}.} \bibinfo{year}{2024}\natexlab{}.
\newblock \showarticletitle{ScanTalk: 3D Talking Heads from Unregistered Scans}. In \bibinfo{booktitle}{\emph{Proceedings of the European Conference on Computer Vision (ECCV)}}.
\newblock


\bibitem[Paszke et~al\mbox{.}(2017)]%
        {pytorch2017}
\bibfield{author}{\bibinfo{person}{Adam Paszke}, \bibinfo{person}{Sam Gross}, \bibinfo{person}{Soumith Chintala}, \bibinfo{person}{Gregory Chanan}, \bibinfo{person}{Edward Yang}, \bibinfo{person}{Zachary DeVito}, \bibinfo{person}{Zeming Lin}, \bibinfo{person}{Alban Desmaison}, \bibinfo{person}{Luca Antiga}, {and} \bibinfo{person}{Adam Lerer}.} \bibinfo{year}{2017}\natexlab{}.
\newblock \showarticletitle{Automatic differentiation in pytorch}.
\newblock  (\bibinfo{year}{2017}).
\newblock


\bibitem[Peng et~al\mbox{.}(2023a)]%
        {selftalk2023}
\bibfield{author}{\bibinfo{person}{Ziqiao Peng}, \bibinfo{person}{Yihao Luo}, \bibinfo{person}{Yue Shi}, \bibinfo{person}{Hao Xu}, \bibinfo{person}{Xiangyu Zhu}, \bibinfo{person}{Hongyan Liu}, \bibinfo{person}{Jun He}, {and} \bibinfo{person}{Zhaoxin Fan}.} \bibinfo{year}{2023}\natexlab{a}.
\newblock \showarticletitle{SelfTalk: A Self-Supervised Commutative Training Diagram to Comprehend 3D Talking Faces}. In \bibinfo{booktitle}{\emph{Proceedings of the 31st ACM International Conference on Multimedia}}. \bibinfo{pages}{5292–5301}.
\newblock
\urldef\tempurl%
\url{https://doi.org/10.1145/3581783.3611734}
\showDOI{\tempurl}


\bibitem[Peng et~al\mbox{.}(2023b)]%
        {emotalk2023}
\bibfield{author}{\bibinfo{person}{Ziqiao Peng}, \bibinfo{person}{Haoyu Wu}, \bibinfo{person}{Zhenbo Song}, \bibinfo{person}{Hao Xu}, \bibinfo{person}{Xiangyu Zhu}, \bibinfo{person}{Jun He}, \bibinfo{person}{Hongyan Liu}, {and} \bibinfo{person}{Zhaoxin Fan}.} \bibinfo{year}{2023}\natexlab{b}.
\newblock \showarticletitle{EmoTalk: Speech-Driven Emotional Disentanglement for 3D Face Animation}. In \bibinfo{booktitle}{\emph{Proceedings of the IEEE/CVF International Conference on Computer Vision (ICCV)}}. \bibinfo{pages}{20687--20697}.
\newblock


\bibitem[Richard et~al\mbox{.}(2021)]%
        {richard2021meshtalk}
\bibfield{author}{\bibinfo{person}{Alexander Richard}, \bibinfo{person}{Michael Zollh{\"o}fer}, \bibinfo{person}{Yandong Wen}, \bibinfo{person}{Fernando De~la Torre}, {and} \bibinfo{person}{Yaser Sheikh}.} \bibinfo{year}{2021}\natexlab{}.
\newblock \showarticletitle{Meshtalk: 3d face animation from speech using cross-modality disentanglement}. In \bibinfo{booktitle}{\emph{Proceedings of the IEEE/CVF International Conference on Computer Vision}}. \bibinfo{pages}{1173--1182}.
\newblock


\bibitem[Sharp et~al\mbox{.}(2022)]%
        {sharp2022diffusionnet}
\bibfield{author}{\bibinfo{person}{Nicholas Sharp}, \bibinfo{person}{Souhaib Attaiki}, \bibinfo{person}{Keenan Crane}, {and} \bibinfo{person}{Maks Ovsjanikov}.} \bibinfo{year}{2022}\natexlab{}.
\newblock \showarticletitle{Diffusionnet: Discretization agnostic learning on surfaces}.
\newblock \bibinfo{journal}{\emph{ACM Transactions on Graphics (TOG)}} \bibinfo{volume}{41}, \bibinfo{number}{3} (\bibinfo{year}{2022}), \bibinfo{pages}{1--16}.
\newblock


\bibitem[Stan et~al\mbox{.}(2023)]%
        {facediffuser2023}
\bibfield{author}{\bibinfo{person}{Stefan Stan}, \bibinfo{person}{Kazi~Injamamul Haque}, {and} \bibinfo{person}{Zerrin Yumak}.} \bibinfo{year}{2023}\natexlab{}.
\newblock \showarticletitle{Facediffuser: Speech-driven 3d facial animation synthesis using diffusion}. In \bibinfo{booktitle}{\emph{Proceedings of the 16th ACM SIGGRAPH Conference on Motion, Interaction and Games}}. \bibinfo{pages}{1--11}.
\newblock
\urldef\tempurl%
\url{https://doi.org/10.1145/3623264.3624447}
\showDOI{\tempurl}


\bibitem[Sun et~al\mbox{.}(2024)]%
        {diffposetalk2024}
\bibfield{author}{\bibinfo{person}{Zhiyao Sun}, \bibinfo{person}{Tian Lv}, \bibinfo{person}{Sheng Ye}, \bibinfo{person}{Matthieu Lin}, \bibinfo{person}{Jenny Sheng}, \bibinfo{person}{Yu-Hui Wen}, \bibinfo{person}{Minjing Yu}, {and} \bibinfo{person}{Yong-Jin Liu}.} \bibinfo{year}{2024}\natexlab{}.
\newblock \showarticletitle{DiffPoseTalk: Speech-Driven Stylistic 3D Facial Animation and Head Pose Generation via Diffusion Models}.
\newblock \bibinfo{journal}{\emph{ACM Transactions on Graphics (TOG)}} \bibinfo{volume}{43}, \bibinfo{number}{4}, Article \bibinfo{articleno}{46} (\bibinfo{year}{2024}), \bibinfo{numpages}{9}~pages.
\newblock
\urldef\tempurl%
\url{https://doi.org/10.1145/3658221}
\showDOI{\tempurl}


\bibitem[Sung-Bin et~al\mbox{.}(2024)]%
        {multitalk2024}
\bibfield{author}{\bibinfo{person}{Kim Sung-Bin}, \bibinfo{person}{Lee Chae-Yeon}, \bibinfo{person}{Gihun Son}, \bibinfo{person}{Oh Hyun-Bin}, \bibinfo{person}{Janghoon Ju}, \bibinfo{person}{Suekyeong Nam}, {and} \bibinfo{person}{Tae-Hyun Oh}.} \bibinfo{year}{2024}\natexlab{}.
\newblock \showarticletitle{MultiTalk: Enhancing 3D Talking Head Generation Across Languages with Multilingual Video Dataset}. In \bibinfo{booktitle}{\emph{Interspeech 2024}}. \bibinfo{pages}{1380--1384}.
\newblock
\showISSN{2958-1796}
\urldef\tempurl%
\url{https://doi.org/10.21437/Interspeech.2024-1794}
\showDOI{\tempurl}


\bibitem[Tan et~al\mbox{.}(2024)]%
        {tan2024say}
\bibfield{author}{\bibinfo{person}{Shuai Tan}, \bibinfo{person}{Bin Ji}, \bibinfo{person}{Yu Ding}, {and} \bibinfo{person}{Ye Pan}.} \bibinfo{year}{2024}\natexlab{}.
\newblock \showarticletitle{Say Anything with Any Style}. In \bibinfo{booktitle}{\emph{Proceedings of the AAAI Conference on Artificial Intelligence}}, Vol.~\bibinfo{volume}{38}. \bibinfo{pages}{5088--5096}.
\newblock


\bibitem[Taylor et~al\mbox{.}(2012)]%
        {taylor2012dynamic}
\bibfield{author}{\bibinfo{person}{Sarah~L Taylor}, \bibinfo{person}{Moshe Mahler}, \bibinfo{person}{Barry-John Theobald}, {and} \bibinfo{person}{Iain Matthews}.} \bibinfo{year}{2012}\natexlab{}.
\newblock \showarticletitle{Dynamic units of visual speech}. In \bibinfo{booktitle}{\emph{Proceedings of the 11th ACM SIGGRAPH/Eurographics conference on Computer Animation}}. \bibinfo{pages}{275--284}.
\newblock


\bibitem[Team et~al\mbox{.}(2023)]%
        {team2023gemini}
\bibfield{author}{\bibinfo{person}{Gemini Team}, \bibinfo{person}{Rohan Anil}, \bibinfo{person}{Sebastian Borgeaud}, \bibinfo{person}{Jean-Baptiste Alayrac}, \bibinfo{person}{Jiahui Yu}, \bibinfo{person}{Radu Soricut}, \bibinfo{person}{Johan Schalkwyk}, \bibinfo{person}{Andrew~M Dai}, \bibinfo{person}{Anja Hauth}, \bibinfo{person}{Katie Millican}, {et~al\mbox{.}}} \bibinfo{year}{2023}\natexlab{}.
\newblock \showarticletitle{Gemini: a family of highly capable multimodal models}.
\newblock \bibinfo{journal}{\emph{arXiv preprint arXiv:2312.11805}} (\bibinfo{year}{2023}).
\newblock


\bibitem[Tevet et~al\mbox{.}(2023)]%
        {tevet2023human}
\bibfield{author}{\bibinfo{person}{Guy Tevet}, \bibinfo{person}{Sigal Raab}, \bibinfo{person}{Brian Gordon}, \bibinfo{person}{Yoni Shafir}, \bibinfo{person}{Daniel Cohen-or}, {and} \bibinfo{person}{Amit~Haim Bermano}.} \bibinfo{year}{2023}\natexlab{}.
\newblock \showarticletitle{Human Motion Diffusion Model}. In \bibinfo{booktitle}{\emph{The Eleventh International Conference on Learning Representations}}.
\newblock
\urldef\tempurl%
\url{https://openreview.net/forum?id=SJ1kSyO2jwu}
\showURL{%
\tempurl}


\bibitem[Tian et~al\mbox{.}(2024)]%
        {var2024}
\bibfield{author}{\bibinfo{person}{Keyu Tian}, \bibinfo{person}{Yi Jiang}, \bibinfo{person}{Zehuan Yuan}, \bibinfo{person}{BINGYUE PENG}, {and} \bibinfo{person}{Liwei Wang}.} \bibinfo{year}{2024}\natexlab{}.
\newblock \showarticletitle{Visual Autoregressive Modeling: Scalable Image Generation via Next-Scale Prediction}. In \bibinfo{booktitle}{\emph{The Thirty-eighth Annual Conference on Neural Information Processing Systems}}.
\newblock
\urldef\tempurl%
\url{https://openreview.net/forum?id=gojL67CfS8}
\showURL{%
\tempurl}


\bibitem[van~den Oord et~al\mbox{.}(2017)]%
        {neuraldiscrete2017}
\bibfield{author}{\bibinfo{person}{Aaron van~den Oord}, \bibinfo{person}{Oriol Vinyals}, {and} \bibinfo{person}{Koray Kavukcuoglu}.} \bibinfo{year}{2017}\natexlab{}.
\newblock \showarticletitle{Neural discrete representation learning}. In \bibinfo{booktitle}{\emph{Proceedings of the 31st International Conference on Neural Information Processing Systems}} (Long Beach, California, USA) \emph{(\bibinfo{series}{NIPS'17})}. \bibinfo{publisher}{Curran Associates Inc.}, \bibinfo{address}{Red Hook, NY, USA}, \bibinfo{pages}{6309–6318}.
\newblock
\showISBNx{9781510860964}


\bibitem[Wang et~al\mbox{.}(2020)]%
        {wang2020mead}
\bibfield{author}{\bibinfo{person}{Kaisiyuan Wang}, \bibinfo{person}{Qianyi Wu}, \bibinfo{person}{Linsen Song}, \bibinfo{person}{Zhuoqian Yang}, \bibinfo{person}{Wayne Wu}, \bibinfo{person}{Chen Qian}, \bibinfo{person}{Ran He}, \bibinfo{person}{Yu Qiao}, {and} \bibinfo{person}{Chen~Change Loy}.} \bibinfo{year}{2020}\natexlab{}.
\newblock \showarticletitle{Mead: A large-scale audio-visual dataset for emotional talking-face generation}. In \bibinfo{booktitle}{\emph{European conference on computer vision}}. Springer, \bibinfo{pages}{700--717}.
\newblock


\bibitem[Wu et~al\mbox{.}(2024)]%
        {wu2024mmhead}
\bibfield{author}{\bibinfo{person}{Sijing Wu}, \bibinfo{person}{Yunhao Li}, \bibinfo{person}{Yichao Yan}, \bibinfo{person}{Huiyu Duan}, \bibinfo{person}{Ziwei Liu}, {and} \bibinfo{person}{Guangtao Zhai}.} \bibinfo{year}{2024}\natexlab{}.
\newblock \showarticletitle{MMHead: Towards Fine-grained Multi-modal 3D Facial Animation}. In \bibinfo{booktitle}{\emph{Proceedings of the 32nd ACM International Conference on Multimedia}}. \bibinfo{pages}{7966--7975}.
\newblock


\bibitem[Xie et~al\mbox{.}(2022)]%
        {xie2022vfhq}
\bibfield{author}{\bibinfo{person}{Liangbin Xie}, \bibinfo{person}{Xintao Wang}, \bibinfo{person}{Honglun Zhang}, \bibinfo{person}{Chao Dong}, {and} \bibinfo{person}{Ying Shan}.} \bibinfo{year}{2022}\natexlab{}.
\newblock \showarticletitle{VFHQ: A High-Quality Dataset and Benchmark for Video Face Super-Resolution}. In \bibinfo{booktitle}{\emph{The IEEE Conference on Computer Vision and Pattern Recognition Workshops (CVPRW)}}.
\newblock


\bibitem[Xing et~al\mbox{.}(2023)]%
        {codetalker2023}
\bibfield{author}{\bibinfo{person}{Jinbo Xing}, \bibinfo{person}{Menghan Xia}, \bibinfo{person}{Yuechen Zhang}, \bibinfo{person}{Xiaodong Cun}, \bibinfo{person}{Jue Wang}, {and} \bibinfo{person}{Tien-Tsin Wong}.} \bibinfo{year}{2023}\natexlab{}.
\newblock \showarticletitle{Codetalker: Speech-driven 3d facial animation with discrete motion prior}. In \bibinfo{booktitle}{\emph{Proceedings of the IEEE/CVF Conference on Computer Vision and Pattern Recognition}}. \bibinfo{pages}{12780--12790}.
\newblock


\bibitem[Xu et~al\mbox{.}(2024)]%
        {xu2023gaussianheadavatar}
\bibfield{author}{\bibinfo{person}{Yuelang Xu}, \bibinfo{person}{Benwang Chen}, \bibinfo{person}{Zhe Li}, \bibinfo{person}{Hongwen Zhang}, \bibinfo{person}{Lizhen Wang}, \bibinfo{person}{Zerong Zheng}, {and} \bibinfo{person}{Yebin Liu}.} \bibinfo{year}{2024}\natexlab{}.
\newblock \showarticletitle{Gaussian Head Avatar: Ultra High-fidelity Head Avatar via Dynamic Gaussians}. In \bibinfo{booktitle}{\emph{Proceedings of the IEEE/CVF Conference on Computer Vision and Pattern Recognition (CVPR)}}.
\newblock


\bibitem[Xu et~al\mbox{.}(2013)]%
        {xu2013practical}
\bibfield{author}{\bibinfo{person}{Yuyu Xu}, \bibinfo{person}{Andrew~W Feng}, \bibinfo{person}{Stacy Marsella}, {and} \bibinfo{person}{Ari Shapiro}.} \bibinfo{year}{2013}\natexlab{}.
\newblock \showarticletitle{A practical and configurable lip sync method for games}.
\newblock In \bibinfo{booktitle}{\emph{Proceedings of Motion on Games}}. \bibinfo{pages}{131--140}.
\newblock


\bibitem[Yang et~al\mbox{.}(2024)]%
        {yang2024probabilistic}
\bibfield{author}{\bibinfo{person}{Karren~D Yang}, \bibinfo{person}{Anurag Ranjan}, \bibinfo{person}{Jen-Hao~Rick Chang}, \bibinfo{person}{Raviteja Vemulapalli}, {and} \bibinfo{person}{Oncel Tuzel}.} \bibinfo{year}{2024}\natexlab{}.
\newblock \showarticletitle{Probabilistic Speech-Driven 3D Facial Motion Synthesis: New Benchmarks Methods and Applications}. In \bibinfo{booktitle}{\emph{Proceedings of the IEEE/CVF Conference on Computer Vision and Pattern Recognition}}. \bibinfo{pages}{27294--27303}.
\newblock


\bibitem[Ye et~al\mbox{.}(2024)]%
        {ye2024real3d}
\bibfield{author}{\bibinfo{person}{Zhenhui Ye}, \bibinfo{person}{Tianyun Zhong}, \bibinfo{person}{Yi Ren}, \bibinfo{person}{Jiaqi Yang}, \bibinfo{person}{Weichuang Li}, \bibinfo{person}{Jiawei Huang}, \bibinfo{person}{Ziyue Jiang}, \bibinfo{person}{Jinzheng He}, \bibinfo{person}{Rongjie Huang}, \bibinfo{person}{Jinglin Liu}, {et~al\mbox{.}}} \bibinfo{year}{2024}\natexlab{}.
\newblock \showarticletitle{Real3D-Portrait: One-shot Realistic 3D Talking Portrait Synthesis}.
\newblock \bibinfo{journal}{\emph{arXiv preprint arXiv:2401.08503}} (\bibinfo{year}{2024}).
\newblock


\bibitem[Yi et~al\mbox{.}(2022)]%
        {yi2022predicting}
\bibfield{author}{\bibinfo{person}{Ran Yi}, \bibinfo{person}{Zipeng Ye}, \bibinfo{person}{Zhiyao Sun}, \bibinfo{person}{Juyong Zhang}, \bibinfo{person}{Guoxin Zhang}, \bibinfo{person}{Pengfei Wan}, \bibinfo{person}{Hujun Bao}, {and} \bibinfo{person}{Yong-Jin Liu}.} \bibinfo{year}{2022}\natexlab{}.
\newblock \showarticletitle{Predicting personalized head movement from short video and speech signal}.
\newblock \bibinfo{journal}{\emph{IEEE Transactions on Multimedia}}  \bibinfo{volume}{25} (\bibinfo{year}{2022}), \bibinfo{pages}{6315--6328}.
\newblock


\bibitem[Zhang et~al\mbox{.}(2024)]%
        {zhang2024letstalk}
\bibfield{author}{\bibinfo{person}{Haojie Zhang}, \bibinfo{person}{Zhihao Liang}, \bibinfo{person}{Ruibo Fu}, \bibinfo{person}{Zhengqi Wen}, \bibinfo{person}{Xuefei Liu}, \bibinfo{person}{Chenxing Li}, \bibinfo{person}{Jianhua Tao}, {and} \bibinfo{person}{Yaling Liang}.} \bibinfo{year}{2024}\natexlab{}.
\newblock \showarticletitle{LetsTalk: Latent Diffusion Transformer for Talking Video Synthesis}.
\newblock \bibinfo{journal}{\emph{arXiv preprint arXiv:2411.16748}} (\bibinfo{year}{2024}).
\newblock


\bibitem[Zhang et~al\mbox{.}(2023)]%
        {zhang2023sadtalker}
\bibfield{author}{\bibinfo{person}{Wenxuan Zhang}, \bibinfo{person}{Xiaodong Cun}, \bibinfo{person}{Xuan Wang}, \bibinfo{person}{Yong Zhang}, \bibinfo{person}{Xi Shen}, \bibinfo{person}{Yu Guo}, \bibinfo{person}{Ying Shan}, {and} \bibinfo{person}{Fei Wang}.} \bibinfo{year}{2023}\natexlab{}.
\newblock \showarticletitle{Sadtalker: Learning realistic 3d motion coefficients for stylized audio-driven single image talking face animation}. In \bibinfo{booktitle}{\emph{Proceedings of the IEEE/CVF Conference on Computer Vision and Pattern Recognition}}. \bibinfo{pages}{8652--8661}.
\newblock


\bibitem[Zhou et~al\mbox{.}(2022)]%
        {zhou2022codeformer}
\bibfield{author}{\bibinfo{person}{Shangchen Zhou}, \bibinfo{person}{Kelvin~C.K. Chan}, \bibinfo{person}{Chongyi Li}, {and} \bibinfo{person}{Chen~Change Loy}.} \bibinfo{year}{2022}\natexlab{}.
\newblock \showarticletitle{Towards Robust Blind Face Restoration with Codebook Lookup TransFormer}. In \bibinfo{booktitle}{\emph{NeurIPS}}.
\newblock


\bibitem[Zhu et~al\mbox{.}(2023)]%
        {zhu2023taming}
\bibfield{author}{\bibinfo{person}{Lingting Zhu}, \bibinfo{person}{Xian Liu}, \bibinfo{person}{Xuanyu Liu}, \bibinfo{person}{Rui Qian}, \bibinfo{person}{Ziwei Liu}, {and} \bibinfo{person}{Lequan Yu}.} \bibinfo{year}{2023}\natexlab{}.
\newblock \showarticletitle{Taming Diffusion Models for Audio-Driven Co-Speech Gesture Generation}. In \bibinfo{booktitle}{\emph{Proceedings of the IEEE/CVF Conference on Computer Vision and Pattern Recognition}}. \bibinfo{pages}{10544--10553}.
\newblock


\bibitem[Zhuang et~al\mbox{.}(2024)]%
        {Zhuang2024learn2talk}
\bibfield{author}{\bibinfo{person}{Yixiang Zhuang}, \bibinfo{person}{Baoping Cheng}, \bibinfo{person}{Yao Cheng}, \bibinfo{person}{Yuntao Jin}, \bibinfo{person}{Renshuai Liu}, \bibinfo{person}{Chengyang Li}, \bibinfo{person}{Xuan Cheng}, \bibinfo{person}{Jing Liao}, {and} \bibinfo{person}{Juncong Lin}.} \bibinfo{year}{2024}\natexlab{}.
\newblock \showarticletitle{Learn2Talk: 3D Talking Face Learns from 2D Talking Face}.
\newblock \bibinfo{journal}{\emph{IEEE Transactions on Visualization and Computer Graphics}} (\bibinfo{year}{2024}), \bibinfo{pages}{1--13}.
\newblock
\urldef\tempurl%
\url{https://doi.org/10.1109/TVCG.2024.3476275}
\showDOI{\tempurl}


\bibitem[Zielonka et~al\mbox{.}(2022)]%
        {MICA2022}
\bibfield{author}{\bibinfo{person}{Wojciech Zielonka}, \bibinfo{person}{Timo Bolkart}, {and} \bibinfo{person}{Justus Thies}.} \bibinfo{year}{2022}\natexlab{}.
\newblock \showarticletitle{Towards Metrical Reconstruction of Human Faces}. In \bibinfo{booktitle}{\emph{European Conference on Computer Vision}}. \bibinfo{pages}{20311--20322}.
\newblock


\end{thebibliography}

\clearpage

\section{Reproducibility}
\subsection{More Data Processing Details of VOCASET}
For evaluation, we also re-tracked the FLAME \citep{FLAME2017} parameters of the VOCASET \citep{voca2019} dataset.
This is necessary because the  original data of VOCASET registers head motion directly onto FLAME meshes rather than FLAME parameters, and also introduces details such as hair that cannot be represented by the FLAME parameters.
Since our method and DiffPoseTalk \citep{diffposetalk2024} works in the FLAME parameter space, we reprocessed VOCASET accordingly. 
Specifically, we obtained the original VOCASET videos, resampled them to 25 fps, and used EMICA \citep{DECA2021, MICA2022, EMOCA2021} to extract a 100-dimensional shape vector, a 50-dimensional expression vector, and a 6-dimensional pose vector for evaluation.  

For mesh-based baseline methods, we used meshes reconstructed from FLAME parameters instead of the original VOCASET meshes.
While this introduces minor differences—such as the absence of hair and subtle expression details—these variations have minimal impact on our primary evaluation, which focuses on lip movements.

\subsection{More Implementation Details}
Our model consists of two main components: a multi-scale encoder-decoder and an autoregressive module.  

The multi-scale encoder-decoder adopts a transformer-based architecture.
Encoder and decoder both consist of 8 layers, each with 8 attention heads and a hidden dimension of 512.
The encoder outputs a 64-dimensional codebook representation, while the decoder reconstructs the original expression and pose parameters.  
During multi-scale quantization, an additional fully connected layer is used in the upsampling process to recover lost information, maintaining an input and output dimension of 64.
The codebook consists of 256 vectors, each with a dimensionality of 64.  

The autoregressive module is a transformer-based autoregressive model.
It has an input and hidden dimension of 768, a conditional vector dimension of 768, and a depth of 12 layers with 12 attention heads.
At the input stage, both the codebook features and the audio encoding features are projected to a 768-dimensional space.
The audio conditions are incorporated into the transformer layers using adaptive instance normalization (AdaIN) \citep{huang2017arbitrary}.  
After generation through the transformer, a fully connected layer maps the output to a 256-dimensional space, followed by a softmax function to compute the probability distribution over the codebook entries.  

We use the AdamW \citep{loshchilov2017decoupled} optimizer with an initial learning rate of \(1 \times 10^{-4}\), trained for 50,000 iterations.
A linear decay learning rate scheduler is applied, gradually reducing the learning rate to \(1 \times 10^{-5}\) over the course of training.

\subsection{More Evaluation Details}
We generally follow the approach of CodeTalker \citep{codetalker2023} to compute LVE and FFD, and additionally compute the mean mouth opening distance (MMD).
Since existing works rarely specify the exact selection of mouth and upper face vertices, we adopt the official FLAME region masks.
\footnote{FLAME\_masks.pkl downloaded from http://s2017.siggraph.org.}
Specifically, the lip/mouth region consists of 254 vertices from the “lips” area, while the upper face region includes 884 vertices from the “eye\_region” and “forehead” areas.

\begin{table}[ht]
    \caption{Different audio encoder performance on TFHP dataset.}
    \label{tab:supp_more_hp}
    \tablestyle{6pt}{1.1}
    \begin{center}
    \scalebox{0.98}{
    \begin{tabular}{lccc}
    \toprule
    Method & LVE$\downarrow$ & FFD$\downarrow$ & MOD $\downarrow$ \\
    \midrule
    Motion Clip Length 100 Layer 3 &  11.03 & 20.82 & 2.01 \\
    Motion Clip Length 100 Layer 4 &  9.58 & 17.26 & 1.89 \\
    Motion Clip Length 100 Layer 6 &  9.81 & 18.46 & 1.92 \\
    Motion Clip Length 150 Layer 5 &  \textbf{7.98} & \textbf{14.09} & \textbf{1.66} \\
    \midrule
    Motion Clip Length 100 Layer 5 &  9.34 & 18.15 & 1.81 \\
    \bottomrule
    \end{tabular}
    }
    \end{center}
\end{table}

\begin{table}[ht]
\caption{Different audio encoder performance on TFHP dataset.}
\label{tab:supp_mimi}
\tablestyle{6pt}{1.1}
\begin{center}
\scalebox{0.98}{
\begin{tabular}{lccc}
\toprule
Method & LVE$\downarrow$ & FFD$\downarrow$ & MOD $\downarrow$ \\
\midrule
Motion Clip Length 8 + MIMI     & 11.01 & 22.79 & 2.14 \\ 
Motion Clip Length 8 + HuBERT   & 11.73 & 24.89 & 2.25 \\ 
Motion Clip Length 25 + MIMI    & 10.33 & 19.65 & 1.99 \\ 
Motion Clip Length 25 + HuBERT  & 10.20 & 19.00 & 1.97 \\ 
Motion Clip Length 50 + MIMI    & 10.10 & 19.16 & 1.95 \\ 
Motion Clip Length 50 + HuBERT  &  9.78 & 18.03 & 1.89 \\ 
Motion Clip Length 100 + MIMI   &  9.97 & 18.48 & 1.93 \\ 
Motion Clip Length 100 + HuBERT &  \textbf{9.34} & \textbf{18.15} & \textbf{1.81} \\ 
\bottomrule
\end{tabular}
}
\end{center}
\end{table}

\section{Preliminaries of 3DMM}
We leverage the widely used 3D morphable model (3DMM), FLAME \citep{FLAME2017}, known for its geometric accuracy and versatility.
Due to its realistic rendering capabilities and flexibility, FLAME has been widely adopted in applications such as facial animation, avatar creation, and facial recognition.
We use FLAME as our representation for facial motion, construct a multi-scale codebook using FLAME parameters and learn speech-driven autoregression within this codebook, effectively leveraging the priors embedded in FLAME. 
This approach provides two key advantages: (1) it reduces the high-dimensional complexity of modeling a large number of mesh vertices, and (2) it enables seamless integration with downstream tasks that utilize FLAME-based representations \citep{gpavatar2024, deng2024portrait4d2, chu2024gagavatar, xu2023gaussianheadavatar}.

The FLAME model represents the head shape as follows:
\begin{equation}
\label{eq:flame}
    TP(\hat{\beta}, \hat{\theta}, \hat{\psi}) = \bar{T} + BS(\hat{\beta};S) + BP(\hat{\theta};P) + BE(\hat{\psi};E),
\end{equation}
where $\bar{T}$ represents the template head avatar mesh, $BS(\hat{\beta};S)$ is the shape blend-shape function to account for identity-related shape variation, $BP(\hat{\theta};P)$ models jaw and neck pose to correct deformations that cannot be fully explained by linear blend skinning, and the expression blend-shapes $BE(\hat{\psi};E)$ capture facial expressions such as eye closure and smiling.

\begin{table}[ht]
\caption{VQ Autoencoder performance on TFHP dataset.}
\label{tab:supp_vq}
\tablestyle{10pt}{1.1}
\begin{center}
\scalebox{0.98}{
\begin{tabular}{lccc}
\toprule
Method & LVE$\downarrow$ & FFD$\downarrow$ & MOD $\downarrow$ \\
\midrule
Motion Clip Length 8    & 1.96 & 8.81 & 0.43 \\ 
Motion Clip Length 25   & 1.97 & 9.57 & 0.44 \\ 
Motion Clip Length 50   & 2.19 & 9.54 & 0.47 \\ 
Motion Clip Length 100  & 2.08 & 8.97 & 0.45 \\ 
\bottomrule
\end{tabular}
}
\end{center}
\end{table}

\begin{table}[t]
\caption{
    Quantitative results on the VFHQ \citep{xie2022vfhq} dataset. 
}
\label{tab:supp_vfhq}
\tablestyle{7pt}{1.05}
\begin{center}
\scalebox{0.95}{
    \begin{tabular}{lccc}
    \toprule
    Method & LVE $\downarrow$ & FFD $\downarrow$ & MOD $\downarrow$ \\
    \midrule
    FaceFormer \citep{faceformer2022}           &  8.07	& 21.65	& 1.94  \\
    CodeTalker \citep{codetalker2023}           &  7.80 & 23.42 & 1.74  \\
    SelfTalk \citep{selftalk2023}               &  7.59 & 28.23 & 1.71  \\
    FaceDiffuser \citep{facediffuser2023}       &  7.63 & 24.61 & 1.78  \\
    MultiTalk$^{*}$ \citep{multitalk2024}       & 10.91 & 31.79 & 2.56  \\
    ScanTalk \citep{scantalk2024}               &  7.77 & 25.26 & 1.79  \\
    UniTalker \citep{fan2024unitalker}          &  8.94 & 26.89 & 2.31  \\
    DiffPoseTalk$^{*}$ \citep{diffposetalk2024} &  7.79 & 23.42 & 1.73  \\
    \midrule
    ARTalk (Ours)$^{*}$                           & \au{6.92} & \au{21.19} & \au{1.53} \\
    \bottomrule
    \end{tabular}
}
\end{center}
\end{table}

\section{More Discussions on ARTalk}
\subsection{Multi-scale Layers and Time Windows}
Here, we provide an extended exploration of the number of layers and time window lengths.
The quantitative results are shown in the Table. \ref{tab:supp_more_hp}.

We use [1, 50, 100] for the 100 motion length with 3 layers, [1, 20, 60, 100] for 100 motion length with 4 layers, [1, 5, 25, 50, 75, 100] for 100 motion length with 6 layers and [1, 20, 50, 100, 150] for 150 motion length with 5 layers.
The baseline version, as described in our main paper, is a 100-frame window with 5 layers: [1, 5, 25, 50, 100].

Our observations indicate that different numbers of multi-scale layers have some impact on performance, but this effect is not particularly significant when the number of layers is similar.
A noticeable impact on performance is only observed when the number of layers is significantly reduced.

In contrast, the window length continues to show a clear positive correlation with performance; that is, longer time windows generally yield better results. 
However, it is important to note that overly long time windows are highly impractical for streaming applications, as the entire window must be collected before generation can begin.
We believe that a 100-frame (4-second) window length strikes a good balance for this purpose.
And the time window can be further reduced when necessary.

\subsection{The Choice of Audio Encoder}
The performance of the audio encoder has a significant impact on the overall model.
In this work, we use the HuBERT \citep{hsu2021hubert} encoder for experiments.
However, HuBERT is primarily designed for longer audio comprehension, which may not be ideal for our windowed input setting.
For instance, when the input is as short as 8 frames (320 ms), HuBERT struggles to effectively capture phonetic features.  

To address this, we experiment with Mimi \citep{kyutai2024moshi}, an audio encoder better suited for streaming input and short audio segments.
As shown in Table \ref{tab:supp_mimi}, Mimi outperforms HuBERT when the input length is short, whereas HuBERT performs better on longer audio segments.
We attribute this to Mimi’s design, which prioritizes capturing fine-grained phonetic details in short audio sequences.
However, its lower feature dimensionality limits its representational capacity for longer segments compared to HuBERT.  
Ultimately, the selection of the audio encoder should be guided by the specific requirements of the application, balancing the need for short-term phonetic precision and long-term representational capacity.

\begin{table}[t]
\caption{
    Quantitative results on the MultiTalk \citep{multitalk2024} dataset. 
}
\label{tab:supp_multi}
\tablestyle{7pt}{1.05}
\begin{center}
\scalebox{0.95}{
    \begin{tabular}{lccc}
    \toprule
    Method & LVE $\downarrow$ & FFD $\downarrow$ & MOD $\downarrow$ \\
    \midrule
    FaceFormer \citep{faceformer2022}           &  9.90	& \au{21.29}	& 2.31  \\
    CodeTalker \citep{codetalker2023}           &  9.59 & 23.50 & 2.08  \\
    SelfTalk \citep{selftalk2023}               &  9.49 & 29.54 & 2.06  \\
    FaceDiffuser \citep{facediffuser2023}       &  9.32 & 25.61 & 2.04  \\
    MultiTalk$^{*}$ \citep{multitalk2024}       &  9.09 & 26.98 & 2.03  \\
    ScanTalk \citep{scantalk2024}               &  9.31 & 26.80 & 2.42  \\
    UniTalker \citep{fan2024unitalker}          & 10.79 & 27.39 & 2.65  \\
    DiffPoseTalk$^{*}$ \citep{diffposetalk2024} & 10.93 & 25.73 & 2.44  \\
    \midrule
    ARTalk (Ours)$^{*}$                           & \au{7.40} & 23.78 & \au{1.64} \\
    \bottomrule
    \end{tabular}
}
\end{center}
\end{table}

\begin{table}[t]
\caption{
    Quantitative results on the MEAD \citep{wang2020mead} dataset. 
}
\label{tab:supp_mead}
\tablestyle{7pt}{1.05}
\begin{center}
\scalebox{0.95}{
    \begin{tabular}{lccc}
    \toprule
    Method & LVE $\downarrow$ & FFD $\downarrow$ & MOD $\downarrow$ \\
    \midrule
    FaceFormer \citep{faceformer2022}           & 12.80	& 30.67	& 2.94  \\
    CodeTalker \citep{codetalker2023}           & 11.36 & 30.90 & 2.59  \\
    SelfTalk \citep{selftalk2023}               & 11.50 & 32.44 & 2.55  \\
    FaceDiffuser \citep{facediffuser2023}       & 10.23 & 31.61 & 2.38  \\
    MultiTalk$^{*}$ \citep{multitalk2024}       & 11.53 & 34.03 & 2.62  \\
    ScanTalk \citep{scantalk2024}               &  9.83 & \au{24.33} & 2.85  \\
    UniTalker \citep{fan2024unitalker}          & 13.72 & 34.18 & 3.39  \\
    DiffPoseTalk$^{*}$ \citep{diffposetalk2024} & 10.28 & 29.40 & 2.43  \\
    \midrule
    ARTalk (Ours)$^{*}$                           & \au{8.64} & 26.33 & \au{1.89} \\
    \bottomrule
    \end{tabular}
}
\end{center}
\end{table}

\subsection{Performance of Multi-scale VQ Autoencoder}
The overall performance of the model is inherently influenced by the VQ autoencoder, which serves as the upper bound for the final model's performance. 
Here in Table \ref{tab:supp_vq}, we present the performance of VQ autoencoder trained with different clip lengths.
We did not observe a significant impact of clip length on the performance of the VQ autoencoder.
For the autoregressive generation results under different segment length settings, please refer to the Table \ref{tab:ablation} in the main paper.

\begin{figure*}[t]
\begin{center}
\centerline{\includegraphics[width=1.0\linewidth]{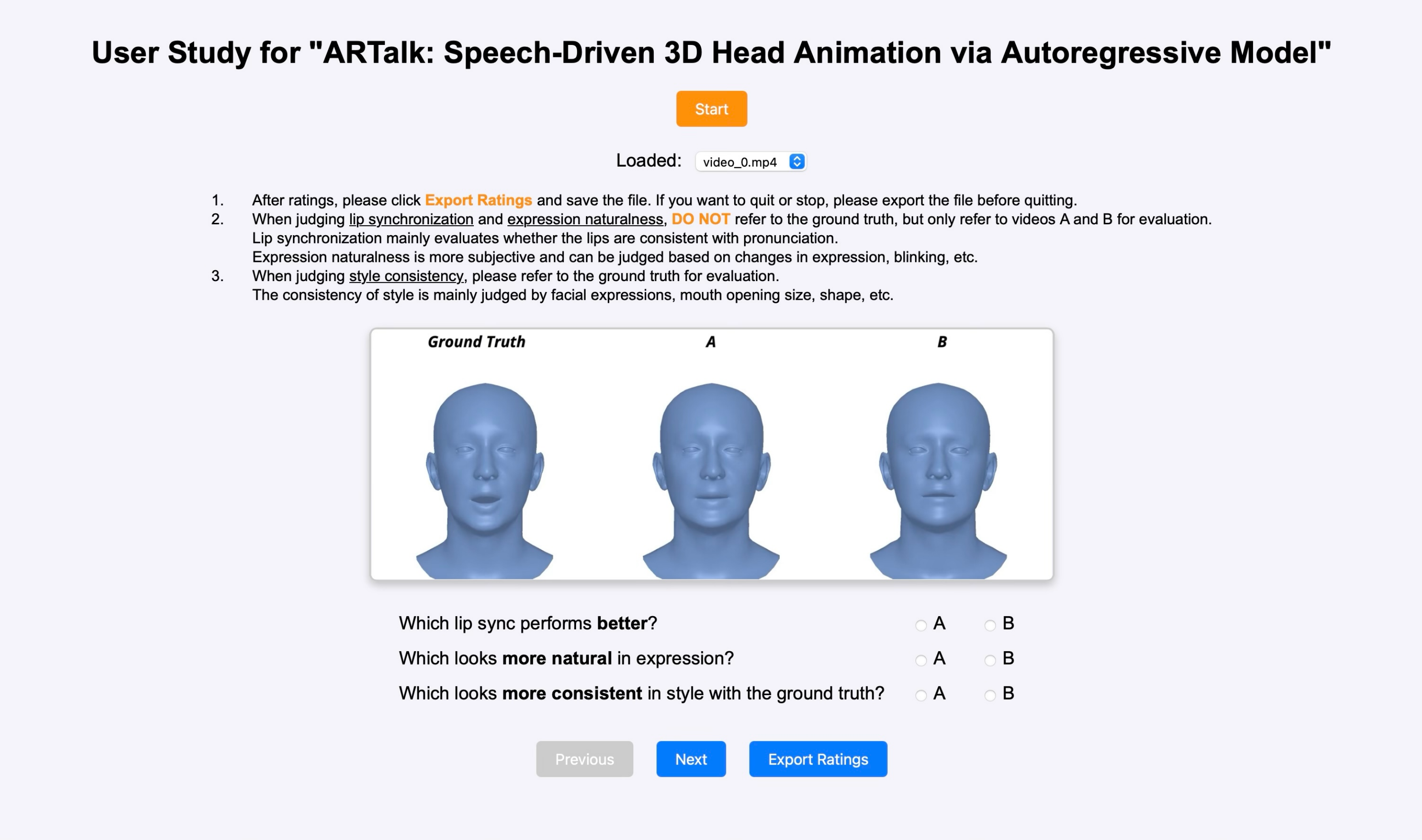}}
\caption{
The interface of our user study.
Users evaluate each video based on three criteria: lip synchronization, expression naturalness, and style consistency.
The first two are judged by comparing method A and B, while style consistency is assessed with reference to the ground truth.
One of the videos (A or B) is generated by our method, and the other by a baseline method, with their order randomized.
}
\label{fig:supp_user_study}
\end{center}
\vskip -0.5cm
\end{figure*}

\section{User Study Details}
We collected a total of 28 survey responses, with each participant answering 96 questions (corresponding to 84 pairwise comparison trials), for a total of 2688 comparisons.
Among these, 63 comparisons were conducted on the TFHP dataset and 21 on the VOCASET dataset.
To mitigate potential biases caused by random or preferential selections, we randomized the display order of our method and the baseline in each trial.
Which means in each comparison, one of the videos (A or B) was generated by our method, but the assignment was randomized.

For each comparison, users were asked three questions.
The first two questions follow prior studies \citep{faceformer2022, codetalker2023}, assessing the quality of lip synchronization and the perceived naturalness of expressions.
Additionally, we introduce a third question to evaluate the consistency of the generated animation with the ground truth style.
The questions presented to users were as follows: Which lip sync performs better? Which looks more natural in expression? Which looks more consistent in style with the ground truth? All three questions are single-choice, requiring users to select either A or B.

\section{More Quantitative Results}
We also provide quantitative results on more datasets in Table \ref{tab:supp_vfhq}, \ref{tab:supp_multi} and \ref{tab:supp_mead}.
Specifically, we have tracked the test sets of VFHQ \citep{xie2022vfhq}, the MultiTalk English subset \citep{multitalk2024}, and MEAD (ID M003) \citep{wang2020mead}, and we have evaluated various methods on these datasets.
For VFHQ, MultiTalk and MEAD, we use models trained on TFHP and only evaluate on these datasets.

\end{document}